\documentclass[conference]{IEEEtran}
\usepackage{amsthm}
\usepackage{graphicx}
\usepackage{supertabular}
\usepackage{slashbox}
\usepackage{subfigure}
\usepackage{rotating}
\usepackage{amssymb}
\usepackage{bm}

\usepackage{epstopdf}
\usepackage{stmaryrd}
\usepackage{amsfonts}
\usepackage{array}
\usepackage{amssymb}
\usepackage{algorithmic}
\usepackage{algorithm}
\usepackage{multirow}
\usepackage[cmex10]{amsmath}
\usepackage{bm}
\usepackage{cite}
\usepackage{color}
\usepackage{threeparttable}
\usepackage{enumerate}
\usepackage{mathrsfs}

\usepackage{tabularx}
\usepackage{makecell, rotating}

\usepackage{setspace}
\usepackage{chngpage}
\usepackage{graphics}
\usepackage{graphicx}
\usepackage{epsfig}

\long\def\comment#1{}
\def\x{\mathbf{x}}
\def\y{\mathbf{y}}

\usepackage{url}

\ifCLASSINFOpdf
\else
\fi
\begin{document}

%
\title{Locally Imposing Function for Generalized Constraint Neural Networks\\
- A Study on Equality Constraints
}

{
\author{\IEEEauthorblockN{Linlin Cao}
\IEEEauthorblockA{NLPR, Institute of Automation\\Chinese Academy of Sciences\\
Beijing, China\\
Email: linlincao\_nlpr@$163$.com}
\and
\IEEEauthorblockN{Ran He}
\IEEEauthorblockA{NLPR, Institute of Automation\\Chinese Academy of Sciences\\
Beijing, China\\
Email: rhe@nlpr.ia.ac.cn}
\and
\IEEEauthorblockN{Bao-Gang Hu}
\IEEEauthorblockA{NLPR, Institute of Automation\\ Chinese Academy of Sciences\\
Beijing, China\\
Email: hubg@nlpr.ia.ac.cn}
}

}


%


\maketitle

\begin{abstract}
This work is a further study on the Generalized Constraint Neural Network (GCNN) model \cite{hu2009a,cao2015generalized}. Two challenges are encountered in the study, that is, to embed any type of prior information and to select its imposing schemes. The work focuses on the second challenge and studies a new constraint imposing scheme for equality constraints. A new method called locally imposing function (LIF) is proposed to provide a local correction to the GCNN prediction function, which therefore falls within Locally Imposing Scheme (LIS). In comparison, the conventional Lagrange multiplier method is considered as Globally Imposing Scheme (GIS)  because its added constraint term exhibits a global impact to its objective function. Two advantages are gained from LIS over GIS. First, LIS enables constraints to fire locally and explicitly in the domain only where they need on the prediction function. Second, constraints can be implemented within a network setting directly. We attempt to interpret several constraint methods graphically from a viewpoint of the locality principle. Numerical examples confirm the advantages of the proposed method. In solving boundary value problems with Dirichlet and Neumann constraints, the GCNN model with LIF is possible to achieve an exact satisfaction of the constraints.
\end{abstract}



%
\IEEEpeerreviewmaketitle

\section{Introduction}
\label{sec: introduction}

Artificial neural networks (\textbf {ANNs}) have received significant progresses after the proposal of deep learning models \cite{lecun2015deep,Deng, Schmidhuber}. ANNs are formed mainly based on learning from data. Hence, they are considered as {\it data-driven} approach \cite{todorovski2006integrating} with a {\it black-box} limitation \cite{olden2002illuminating}. While this feature provides a flexiblility power to ANNs in modeling, they miss a functioning part for {\it top-down mechanisms}, which seems to be necessary for realizing {\it human-like} machines. Furthermore, the ultimate goal of machine learning study is insight, not machine itself. The current ANNs, including deep learning models, fail to present interpretations about their learning processes as well as the associated physical targets, such as human brains.

For adding transparency to ANNs, we proposed a generalized constraint neural network (\textbf {GCNN}) approach \cite{hu2009a, yang2008structural, qu2011generalized}. It can also be viewed as a {\it knowledge-and-data-driven modeling} (\textbf {KDDM}) approach \cite{ran2014determining, fan2015a} because two submodels are formed and coupled as shown in Fig. 1. To simplify discussions later, we refer GCNN and KDDM approaches to the same model.

GCNN models were developed based on the previously existing modeling approaches, such as the ``hybrid neural network (\textbf {HNN})" model \cite{psichogios1992hybrid,thompson1994modeling}. We chose ``{\it generalized constraint}" as the descriptive terms so that a mathematical meaning is stressed \cite{hu2009a}. The terms of generalized constraint was firstly given by Zadeh in 1990's \cite{zadeh1986outline, zadeh1996fuzzy} for describing a wide variety of constraints, such as probabilistic, fuzzy, rough, and other forms. We consider that the concepts of generalized constraint provide us a critical step to construct human-like machines. Implications behind the concepts are at least two challenges as follows.
\begin{itemize}
\item[1.] How to utilize any type of prior information that holds one or a combination of limitations in modeling \cite{hu2009a}, such as ill-defined or unstructured prior.   %
\item[2.] How to select coupling forms in terms of explicitness \cite{hu2009a, qu2011generalized}, physical interpretations \cite{fan2015a}, performances \cite{qu2011generalized, fan2015a}, locality principle\cite{denning2005locality}, and other related issues. %
\end{itemize}

\begin{figure}[t]
\centering
\includegraphics[scale = 0.8]{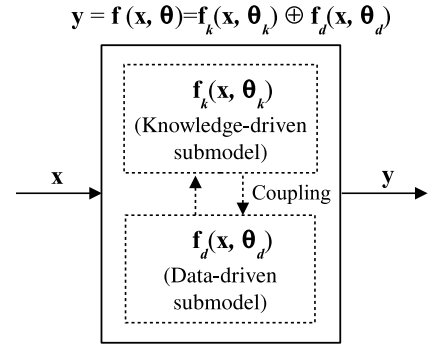}
\caption{Schematic diagram of a KDDM model \cite{ran2014determining, fan2015a}. A GCNN model is formed when the data-driven submodel is ANNs \cite{hu2009a}. }
\label{fig:RBF}
\end{figure}

The first challenge above aims to mimic the behavior of human beings in decision making. Both {\it deduction} and {\it induction} inferences are employed in our daily life. The second challenge attempts to emulate the {\it synaptic plasticity} function of human brain. We are still far away from understanding mathematically how human brain to select and change the couplings. The two challenges lead to a further difficulty as stated in \cite{hu2009a}:
``{\it Confronting the large diversity and unstructured representations of prior knowledge, one would be rather difficult to build a rigorous theoretical framework as already done in the elegant treatments of Bayesian, or Neuro-fuzzy ones}".
The difficulty implies that we need to study GCNN approaches on a class-by-class basis. This work extends our previous study of GCNN models on a class of equality constraints \cite{cao2015generalized}, and focuses on the locality principle in the second challenge. The main progress of this work is twofold below.
\begin{itemize}
\item[1.] A novel proposal of ``Locally Imposing Scheme (\textbf {LIS})'' is presented, resulting in an alternative solution different from ``Globally Imposing Scheme (\textbf {GIS})", such as Lagrange multiplier method.   %
\item[2.] Numerical examples are shown for a class of equality constraints including a derivative form and confirm the specific advantages of LIS over GIS on the given examples. %
\end{itemize}
We will limit the study on the regression problems with equality constraints. The remaining of the paper is organized as follows. Section \ref{sec: Problem} discusses the differences between machine learning problems and
optimization problems. Based on the discussions, the main idea behind LIS is presented. The conventional RBFNN model and its learning are briefly
introduced in Section \ref{sec: RBF}. Section \ref{sec: LIF} demonstrates the proposed
model and its learning process. Numerical experiments on two synthetic
data sets are presented in Section \ref{sec: experiments}. Discussions of locality principle and coupling forms 
are given in Section \ref{sec:Discussions}. Section \ref{sec:remarks} presents final remarks about the work.

\section{Problem discussions and main idea}
\label{sec: Problem}
Mathematically, machine learning problems can be equivalent to optimization problems.
We will compare them for reflecting their differences.
An optimization problem with equality constraints is expressed in the following form \cite{boyd2004convex}:
\begin{eqnarray}
\begin{array}{ll}
 \min ~ F(\textbf{x})  \\
 s.t. ~   G_i(\textbf{x}) = 0, ~ i= 1, 2, \cdots
\end{array}
\label{eq:1}
\end{eqnarray}
where $F(\textbf{x}): R^{d} \rightarrow R$, is the objective function to be minimized over the variable $\textbf{x}$,
and $G_i(\textbf{x})$ is the {\it i}th equality constraint.
In machine learning, its problem having equality constraints can be formulated as \cite{bishop2006pattern}:
\begin{eqnarray}
\begin{array}{ll}
 \min ~ \mathbb{E}[(y - f(\textbf{x}))^2],   \\
 s.t. ~   g_i(\textbf{x}) = 0, ~ i= 1, 2, \cdots
\end{array}
\label{eq:2}
\end{eqnarray}
where $\mathbb{E}$ is an expectation, $f(\textbf{x}): R^{d} \rightarrow R$, is the prediction function which can be formed from
a composition of radical basis functions (\textbf {RBFs}),
and $g_i(\textbf{x})$ is the {\it i}th equality constraint.

Eq. (2) presents several differences in comparing with Eq. (1). For a better understanding, we explain them by imaging a 3D mountain (or a two-input-single-output model). First, while a conventional optimization problem is to search for an optimization point on a {\it well-defined mountain} (or objective function $F$), a machine learning problem tries to form an {\it unknown mountain} (or prediction function $f$) with a minimum error from the observation data. Second, the equality constraints in the optimizations imply that the solution should be located at the constraints. Otherwise, there exist no feasible solutions. For a machine learning problem, the equality constraints suggest that an unknown mountain (or prediction function) surface should go through the given form(s) described by function(s) and/or value(s). If not, an approximation should be made in a minimum error sense. Third, machine learning produces a larger variety of constraint types which are not encountered in the conventional optimization problems.
The main reason is that $g_i(\textbf{x})$ comes from a prior to describe the unknown real-system function.
Sometimes, $g_i(\textbf{x})$ is not well defined, but only shows a ``partially known relationship (\textbf {PKR})" \cite{hu2009a}.
This is why the terms of generalized constraints are used in the machine learning problems. For this reason, we rewrite (2)
in a new form from \cite{hu2009a} to highlight the meaning of $g_i(\textbf{x})$ in the machine learning problems:
\begin{eqnarray}
\begin{array}{ll}
 \min ~ \mathbb{E}[(y - f(\textbf{x}))^2],   \\
 s.t. ~   R_i \langle f \rangle= g_i(\textbf{x}) = 0, ~ \textbf{x} \in C_i, ~i= 1, 2, \cdots
\end{array}
\label{eq:3}
\end{eqnarray}
where $R_i \langle f \rangle$ is the {\it i}th partially known relationship about the function $f$, and $C_i$ is the {\it i}th constraint set for $\textbf{x}$.

	Based on the discussions above, we present a new proposal, namely ``Locally Imposing Scheme (\textbf {LIS})'', in dealing with the equality constraints in machine learning problems. The main idea behind the LIS is realized by the following steps.
	
	Step 1.  The modified prediction function, say, $F(\textbf{x})$, is formed by two basic terms. The first is an original prediction function from unconstrained learning model and the second is the constraint functions $g_i(\textbf{x})$.
	
	Step 2. When the input \textbf{x} is located within the constraint set $C_i$, one enforces $F(\textbf{x})$ to satisfy the function $g_i(\textbf{x})$. Otherwise, $F(\textbf{x})$ is approximately formed from all data excepted for those data within constraint sets.
	
	Step 3. For removing the jump switching in Step 2, we use ``Locally Imposing Function (\textbf {LIF})" as a weight on the constraint term and the complementary weight on the first term so that a continuity property can be held to the modified prediction function $F(\textbf{x})$.
	
	The idea of the first two steps have been reported from the previous studies, particularly in boundary value problems (\textbf{BVPs})  \cite{lagaris1998artificial,hong2009new,mcfall2009artificial}. They used different methods to realize Step 2, such as polynomial methods in \cite{lagaris1998artificial}, RBF methods in \cite{hong2009new}, and length methods in \cite{mcfall2009artificial}. If equality constraints are given by interpolation points, other methods are shown \cite{lauer2008incorporating, hu2009a,qu2011generalized}. Hu, et al. \cite{hu2009a} suggested that ``{\it neural-network-based models can be enhanced by integrating them with the conventional approximation tools}". They showed an example to realize Step 2 and apply Lagrange interpolation method. In the following-up study, an elimination method was used in \cite{qu2011generalized}. All above methods, in fact, fall into the GIS category. In \cite{cao2015generalized}, Cao and Hu applied the LIF method to realize Step 2 and demonstrated that equality function constraints are satisfied completely and exactly on the given Dirichlet boundary (see Fig 4(e) in \cite{cao2015generalized}) but the LIF was not smooth in that work.

	We can observe that the LIS is significantly different from the conventional Lagrange multiplier method that belongs to ``Globally Imposing Scheme (\textbf {GIS)}" because the Lagrange multiplier term exhibits a global impact on an objective function. A heuristic justification for the use of  the LIS is an analogy to the locality principle in the brain functioning of memory \cite{denning2005locality}. All constraints can be viewed as memory. The principle provides both {\it time efficiency} and {\it energy efficiency}, which implies that constraints are better to be imposed through a local means. The LIS in together with the GIS will open a new direction to study the coupling forms towards {\it brain-inspired} machines.

 \section{Conventional RBF neural networks}
\label{sec: RBF}

Given the training data set $X = [\textbf{x}_1, \ldots, \textbf{x}_n]^T$ and its desired outputs $\textbf{y} = [y_1, \ldots, y_n]^T$, where $\textbf{x}_i\in R^{1 \times d}$ is an input vector, $y_i\in R$ denotes the vector of desired network output for the input $\textbf{x}_i$ and $n$ is the number of training data. The output of \textbf{RBFNN} is calculated according to
\begin{eqnarray}
 f(\textbf{x}_i) &=& \sum_{j=1}^m w_j \cdot \phi_j(\textbf{x}_i) = \Phi(\textbf{x}_i) W,
 \label{eq: rbf output function}
 \\
 \phi_j(\textbf{x}_i) &=& \exp(-\|\textbf{x}_i-\boldsymbol{\mu}_j\|^{2}/\sigma_{j}^{2}),
 \label{eq: brf feature mapping function}
\end{eqnarray}
where $W = [w_0, w_1, \ldots, w_m]^T \in R^{(m+1) \times 1}$ represents the model parameter, and $m$ is the number of neurons of the hidden layer. In terms of the feature mapping function $\Phi(X) = [1, \phi_1(X), \ldots, \phi_m(X)] \in R^{ n\times(m+1)}$ (for simplicity, it is denoted as $\Phi$ hereafter), both the centers $U = [\boldsymbol{\mu}_1, \ldots, \boldsymbol{\mu}_m]^T \in R^{m \times d}$ and the widths $\boldsymbol{\sigma} = [\sigma_1, \ldots, \sigma_m]^T\in R^{m \times 1}$ can be easily determined using the method proposed in \cite{schwenker2001three}.

A common optimization
criterion is the mean square error between the actual and
desired network outputs. Therefore, the optimal set of weights
minimizes the performances measure:
\begin{equation}
 \arg\min_{W} \ell_2(W) = \sum_{i=1}^n (y_i - f(\textbf{x}_i) )^2
 = \| \textbf{y} - f(X) \|_2^2,
 \label{eq: objective of RBF}
\end{equation}
where $f(X) = [f(\textbf{x}_1), \ldots, f(\textbf{x}_n)]^T\in R^{n \times 1}$ denotes the prediction outputs of RBFNN.

 Least squares algorithm is used in this work, resulting in the following optimal model parameter
 \begin{equation}
 W^* = (\Phi^T \Phi)^+ \Phi^T \y,
 \end{equation}
 where $(\Phi^T \Phi)^+$ denotes the pseudo-inverse of $\Phi^T \Phi$.

\section{GCNN with equality constraints}
\label{sec: LIF}

In this section, we focus on GCNN with equality constraints (called \textbf{GCNN\_EC} model) by using LIF. Note that LIF is a special method
within LIS category that may include several methods. We first describe a locally imposing function used in {GCNN\_EC} models. 
Then {GCNN\_EC} designs from direct and derivative constraints of $f(\textbf{x})$ are discussed respectively. 
For simplifying presentations, we only consider a single constraint in the design so that the process steps are clear on each individual constraint. Multiple sets 
and combinations of direct and derivative constraints can be extended directly.  

\subsection{Locally Imposing Function}

For realizing Step 3 in Section II, we select Cauchy distribution for
the LIF. The Cauchy distribution is given by: 
\begin{flalign}
f(x;x_{0},\gamma)=\frac{1}{\pi\gamma[1+(\frac{x-x_{0}}{\gamma})^2]},
\label{eq:cauchy}
\end{flalign}
where $x_{0}$ is the location parameter which defines the peak of the distribution, $\gamma ~ (>0)$ is a scale parameter which describes the width of the half of the maximum.
The Cauchy distribution is {\it smooth} and has an {\it infinitely differentiable} property. Other smooth function can also be used as LIF. 

In the context of multi-input variables, we define the LIF of GCNN\_EC in a form of:

\begin{flalign}
\Psi(\bold{\Delta};\gamma)=\frac{1}{\pi\gamma[1+(\frac{\bold{\Delta}}{\gamma})^2]\psi_{norm}},
\label{eq:L}
\end{flalign}
where $\bold{\Delta}(\geq 0)$ denotes the distance variable from $\textbf{x}$ to the constraint location. $\psi_{norm}$ is a normalized parameter 
and ensures a normalization on $0<\Psi\leq1$. $\Psi(\bold{\Delta};\gamma)$ is a monotonically decreasing function with respect to the distance $\bold{\Delta}$. 
We call $\gamma$ a {\it locality parameter} because it controls the locality property of the LIF. 
When $\gamma$ decreases, $\Psi$ becomes sharper in its function shape. Generally, we preset this parameter as a constant by a trial and error way.  
Hence, we drop $\gamma$ to describe $\Psi(\bold{\Delta})$.

\subsection{Equality constraints on $f(\textbf{x})$}
Suppose the output of the network strictly satisfies a single equality constraint given by:
\begin{flalign}
f(\textbf{x})  = f_{C}(\textbf{x}), \textbf{x}\in C ,
\label{eq: equality constraint}
\end{flalign}
where $C$ denotes a constraint set for $\textbf{x}$, $ f_{C}$ can be any numerical value or function. 
Note that BVPs with a Dirichlet form are a special case in Eq. (\ref{eq: equality constraint}) because $ f_{C}$ may be given on any regions without a limitation on boundary.
Facing the following constrained minimization problem:
\begin{multline}
~~~~~ ~~~~~~~~\min_{W} \ell_2(W) = \| \textbf{y} - f(X) \|_2^2   \\[-4pt]
 s.t. \ \ \  f(\textbf{x})  = f_{C}(\textbf{x}), \textbf{x}\in C, ~~~~~~~~~~~~~  ~~\
\end{multline}
a conventional RBFNN model generally applies a Lagrange multiplier and transfers it into an unconstrained problem by
\begin{multline}
\min_{W, \lambda} \ell_2(W, \lambda)  = \| \textbf{y} - f(X) \|_2^2    
 +  \lambda (f(\textbf{x} \in C)  - f_{C}(\textbf{x})),  \
\end{multline}
where $\lambda$ is a new variable determined from the above solution. 
Different with Lagrange multiplier method which imposes a constraint in a global manner on the objective function, we use LIS to solve a constrained optimization problem. 
A modified prediction function is defined in {GCNN\_EC} by 
\begin{flalign}
& f_{W,C} (\textbf{x}) = (1-\Psi(\bold{\Delta})) f(\textbf{x}) + \Psi(\bold{\Delta}) f_{C}(\textbf{x}),  
\label{eq: modified output function}
\end{flalign}
so that one solves an unconstrained problem in a form of:

\begin{equation}
 \min_{W} \ell_2(W) = \| \textbf{y}  - f_{W,C}(X) \|_2^2 .
 \label{eq: objective1}
\end{equation}
One can observe that $f_{W,C}(\textbf{x})$ contains two terms. Both terms are associated with the smooth LIF in Eq. (\ref{eq:L}) so that $f_{W,C} (\textbf{x})$ is possible to hold a smoothness property. 
One important relation can be proved directly from Eqs. (\ref{eq:L}) and (\ref{eq: modified output function}):
\begin{flalign}
f_{W,C}(\textbf{x})=f_{C}(\textbf{x}), \textbf{x}\in C, ~when ~ \bold{\Delta}= \bold {0}.
\label{eq: value_constraints}
\end{flalign}
The above equation indicates an exact satisfaction on the constraint for {GCNN\_EC} models.  

In this work, we still follow the way in presenting $\mu_j$ and $\sigma_j$ to RBF models \cite{schwenker2001three,qu2011generalized} and determining only weight parameters $w_j$ from solving a linear problem. Its optimal solution for {GCNN\_EC} is given below:
 \begin{equation}
 \label{eq:gcnnef}
 \begin{split}
 W^* = &[((\textbf{1}-\Psi) \circ \Phi ^T) ((\textbf{1}-\Psi)^T \circ \Phi)]^+ \\
 &[(\textbf{1}-\Psi) \circ \Phi^T] (\y - \Psi(X) \circ \mathbf{f_{C}}),
 \end{split}
 \end{equation}
 where $\circ$ denotes the Hadamard product\cite{horn1990hadamard},
 $\Psi= [\Psi(X), \ldots, \Psi(X)]^T\in R^{(m+1)\times n}$, $\Psi(X)=[\Psi(\textbf{x}_1),\cdots,\Psi(\textbf{x}_n)]^T$ and $\mathbf{f_{C}}= [f_{C}(\textbf{x}_1), \ldots, f_{C}(\textbf{x}_n)]^T $. $\textbf{1}$ is a matrix whose elements are equal to 1 and has the same size as $\Psi$.

\subsection{Equality constraints on derivative of $f(\textbf{x})$}
In BVPs, the constraints with the derivative of $f(\textbf{x})$ are Neumann forms.
Suppose that the output of a RBFNN satisfies a known derivative constraint:
\begin{flalign}
\frac{\partial f(\textbf{x})}{\partial x_{k}}=(f_{C}(\textbf{x}))^1_k, \textbf{x}\in C ,
\label{eq: dao constraint}
\end{flalign}
where the superscript $1$ and the subscript $k$ describe a first order partial differential equation with respect to the {\it k}th input variable for $ f_{C}(\x)$. 
Two cases will occur in designs of {GCNN\_EC} models as shown below.  

\subsubsection{General case: non-integrable to derivative constraints}
\label{sec:gcnnec_u}
A general case is that an explicit form of $f_{C}(\textbf{x})$
cannot be derived from its given Neumann constraint.
A modified loss function, including two terms, is given by the following form  so that the constraint is approximately satisfied as much
as possible:
\begin{multline}
\label{eq: Gen}
\min_{W} \ell_2(W) = (\textbf{1}-\Psi(X))^{T}\circ (\textbf{y} - f(X))^{T}(\textbf{y} - f(X))  +    \\[-2pt]
  \Psi(X)^{T}\circ ((f(\textbf{x} \in C))^1_k-(f_{C}(\textbf{x}))^1_k)^{T}((f(\textbf{x} \in C))^1_k-(f_{C}(\textbf{x}))^1_k). \\ \
\end{multline}
The optimization solution is then given by
 \begin{equation}
 \begin{split}
 W^* =& [(\textbf{1}-\Psi)\circ \Phi^T \Phi +\Psi \circ (\Phi^{1}_k) ^T \Phi^{1}_k]^+ \\
 &[(\textbf{1}-\Psi)\circ \Phi^T \y+\Psi\circ (\Phi^{1}_k)^T\mathbf{f_{C}}],
 \end{split}
 \label{eq: modified equality constraints}
 \end{equation}
where $\Phi^{1}_k=[(\Phi(\textbf{x}_{1}))^1_k, \cdots, (\Phi(\textbf{x}_{n}))^1_k ]^T$. 
The LIS idea behind the loss function in (\ref{eq: Gen}) is not limited to the derivative constraints and is possible to apply for other types of equality constraints.

\subsubsection{Special case: integrable to derivative constraints}
This is a special case because it requires that $f_{C}(\textbf{x})$ should be derived from the given constraints for realizing an explicit form. 
In other words, a Neumann constraint is integrable, 
\begin{eqnarray}
\begin{array}{ll}
f_C(\textbf{x})=\int \frac{\partial f_C(\textbf{x})}{\partial x_{k}}d{x_k}=f^0_C(\textbf{x})+{c},
\end{array}
\label{eq:integral}
\end{eqnarray}
so that an integration term $f^0_C(\textbf{x})$ is exactly known in (\ref{eq:integral}). The above constant $c$ is neglected because {GCNN\_EC} includes this term already.  
Hence, by substituting (\ref{eq:integral}) into (\ref{eq: modified output function}), one will solve a BVP with a Neumann constraint like with a Dirichlet constraint.
However, for distinguishing with the { GCNN\_EC} model in the general case, we denote {\textbf{GCNN\_EC\_I}} model in the special case for a Neumann constraint. 

\section{Numerical examples}
\label{sec: experiments}

Numerical examples are shown in this section for comparisons between LIS and GIS. When GCNN\_EC is a model within LIS, the other models, GCNN + Lagrange, BVC-RBF \cite{hong2009new}, RBFNN + Lagrange interpolation\cite{hu2009a} and GCNN-LP \cite{qu2011generalized}, are considered in GIS.  
\subsection{``Sinc'' function with interpolation point constraints}
The first example is on interpolation point constraints. 
Consider the problem of approximating a $Sinc = sin(x)/x$ function based on the equality constraints $ f(0)=1$ and $f(\pi/2)=2/\pi$. 
The function is corrupted by an additive Gaussian noise $N(0,0.05^2)$.
This optimization problem can be represented as:
 \begin{eqnarray}
 \begin{split}
& \quad \min_{W} \ell_2(W) =\| \y - f(X) \|_2^2 ,\\
& \quad \ s.t. \ \quad f(0)=1,\\
& \ \ \ \quad \quad \quad   f(\pi/2)=2/\pi.
\end{split}
 \end{eqnarray}
 The training data have 30 instances generated uniformly along $x$ variable at the intervals $[-10,10]$, and 500 testing data are randomly generated from the same intervals. 
 Table \ref{tabel_SINC} shows the performances of six methods, in which only RBFNN does not belong to a constraint method. We examine performances on both constraints and all testing data.
One can observe that among the five constraint methods, RBFNN + Lagrange multiplier presents an excellent approximation ($\approx 0.00$) on the constraints, and the others produce an exact satisfaction ($=\textbf{0}$ for an exact zero) on the constraints.

%

\begin{table*}[htbp]
\centering
\scriptsize
\caption{Results for a 'sinc' function with two interpolation-point constraints.($N_{train}$ is the number of training data, $N_{test}$ is the number of testing data, $N_{RBF}$ is the number of RBF. MSE(Mean $\pm$ Standard) means the average and standard deviation on the 100 groups of test data. $MSE\_{cstr}$ is the MSE on the constraints, $MSE\_{test}$ is the MSE on testing data. Additive noise is $N(0,0.05^2)$.)}
\label{tabel_SINC}
\begin{tabular}{|c|c|c|c|c|c|c|c|}
\hline
 Method &$N_{train}$& $N_{test}$ &$N_{RBF}$ & Key parameter(s) & $MSE\_{cstr}(\times10^{-3})$ &$MSE\_{test}(\times10^{-3})$\\
\hline
RBFNN&30&500&$11$ &  & $0.91\pm0.84 $ & $3.81\pm3.70 $  \\
\hline
RBFNN+Lagrange multiplier& 30&500&$11$ &  &$ \approx 0.00\pm0.00$ & $\textbf{3.73}\pm\textbf{3.78}$ \\
\hline
BVC-RBF\cite{hong2009new}&30&500&$11$&$\tau_{1}=\tau_{2}=2$ & $\textbf{0}\pm\textbf{0}$ & $3.82\pm3.73 $   \\
\hline
GCNN+Lagrange interpolation\cite{hu2009a}& 30&500&$11$ &  &$\textbf{0}\pm\textbf{0}$ & $3.83\pm3.74$ \\
\hline
GCNN-LP\cite{qu2011generalized}& 30&500&$11$ & &$\textbf{0}\pm\textbf{0}$ & $ 3.81\pm 3.70 $\\
\hline
GCNN\_EC& 30&500&$11$ & $\gamma=0.0001$ &$\textbf{0}\pm\textbf{0}$ & $3.80\pm3.71$ \\
\hline
\end{tabular}
\end{table*}

\subsection{Partial differential equation(\textbf{PDE}) with a Dirichlet boundary condition}

The boundary value problem \cite{hong2009new} is given by
\begin{eqnarray}
\begin{array}{ll}
 [\frac{\partial^{2}}{\partial x_{1}^{2}}+\frac{\partial^{2}}{\partial x_{2}^{2}}]f(x_{1},x_{2})=e^{-x_{1}}(x_{1}-2+x_{2}^{3}+6x_{2}) \\
 x_{1}\in[0,1],  x_{2}\in[0,1],
\end{array}
\label{PDE}
\end{eqnarray}
with a Dirichlet boundary condition by
\begin{equation}\label{PDE_Di}
f(0,x_{2})=x_{2}^3.
\end{equation}

The analytic solution is
\begin{equation}\label{PDE_solution}
f(x_{1},x_{2})=e^{-x_{1}}(x_{1}+x_{2}^{3}).
\end{equation}

The optimization problem with a Dirichlet boundary is:
\begin{eqnarray}
\begin{split}
& \min_{W} \ell_2(W) = \| \y - f(X) \|_2^2, \\
&  s.t.  \quad f(0,x_{2})=x_{2}^3.
\end{split}
\label{eq:objective of Di}
\end{eqnarray}

\begin{figure}[t]
\centering
\includegraphics[scale = 0.4]{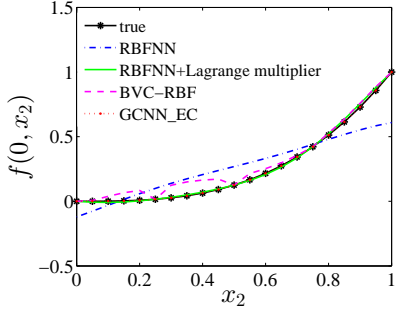}
\caption{Plots on the boundary $(x_{1}=0, x_{2})$ with the Dirichlet constraint.}
\label{PDE_Dirichlet}
\end{figure}

A Gaussian noise $N (0, 0.1^2 )$ is added onto the original function (\ref{PDE_solution}).
The training data have 121 instances selected evenly within $x_{1},x_{2}\in[0,1]$. The testing data have 321 instances, where 300 instances are randomly sampled within $x_{1},x_{2}\in[0,1]$ and 21 instances selected evenly in the boundary (0, $x_{2}$). Because RBFNN+Lagrange multiplier, BVC-RBF, and GCNN+Lagrange interpolation are applicable for solving this problem only after  transferring a {\it ``continuous constrain''} \cite{qu2011generalized} into {\it ``point-wise constraints''}. For this reason, we select 5 points evenly according to (\ref{PDE_Di}) along the boundary (0, $x_{2}$) for them. 
Table \ref{tabel_PDE_Dirichlet} lists the fitting performances in the boundary and the testing data. GCNN\_EC can satisfy the Dirichlet boundary condition exactly for a continuous function constrain. The other constraint methods can reach the satisfaction only on the point-wise constraint location (Fig. \ref{PDE_Dirichlet}). 
Moreover, GCNN\_EC performs much better than the other methods in the testing data.

\begin{table*}[htbp]
\centering
\scriptsize
\renewcommand\arraystretch{1.0}
\caption{Results for a PDE example with the Dirichlet boundary condition.
($N_{train}$ is the number of training data, $N_{test}$ is the number of testing data, $N_{RBF}$ is the number of RBF, $N_{pwc}$ is the number of point-wise constraints along the boundary. MSE(Mean $\pm$ Standard) means the average and standard deviation on the 100 groups of test data. $MSE\_{cstr}$ is the MSE on the constraints, $MSE\_{test}$ is the MSE on testing data. Additive noise is $N(0,0.1^2)$.)}
\label{tabel_PDE_Dirichlet}
\newcommand{\tabincell}[2]{\begin{tabular}{@{}#1@{}}#2\end{tabular}}
\begin{tabular}{|c|c|c|c|c|c|c|c|}
\hline
 Method &$N_{train}$& $N_{test}$ &$N_{RBF}$ &$N_{pwc}$& Key Parameter(s) & $MSE\_{cstr}$ &$MSE\_{test}$\\
\hline
RBFNN&121&321&$10$ &0& & $0.0079\pm0.0043$ & $0.0092\pm0.0091 $  \\
\hline
RBFNN+Lagrange multiplier& 121&321&$10$ &5& &$0.0002\pm0.0001$ & $1.8614\pm4.3791$ \\
\hline
BVC-RBF\cite{hong2009new}&121&321&$10$ &5&$\tau_{1}=\tau_{2}=0.6$& $0.0019\pm0.0014$ & $0.0076\pm0.0087 $   \\
\hline
GCNN\_EC& 121&321&$10$ &0& $\gamma=0.5$ &$\textbf{0}\pm\textbf{0}$ & $\textbf{0.0074}\pm\textbf{0.0087}$ \\
\hline
\end{tabular}
\end{table*}

\subsection{PDE with a Neumann boundary condition}
In this example, the boundary value problem (\ref{PDE}) is given with a Neumann boundary condition by:
\begin{eqnarray}
\begin{split}
& \min_{W} \ell_2(W) = \| \y - f(X) \|_2^2, \\
&  s.t.  \quad \frac{\partial f(x_{1},x_{2})}{\partial x_{2}}|_{x_{1}=0}=3x_{2}^{2}.
\end{split}
\label{eq:objective of Nu}
\end{eqnarray}

No additive noise is added in this case study. Generally, RBFNN+Lagrange multiplier, BVC-RBF, and GCNN+Lagrange interpolation methods fail in this case if without transferring to point-wise constraints. 
We use GCNN\_EC and GCNN\_EC\_I to solve this constraint problem and compare their performances. RBFNN is also given but without using the constraint. 
The training data have 121 instances selected evenly within $x_{1},x_{2}\in[0,1]$. The testing data have 321 instances, where 300 instances are randomly sampled within $x_{1},x_{2}\in[0,1]$ and 21 instances are selected evenly in the boundary (0,$x_{2}$). 

\begin{figure}[t]
\centering
\includegraphics[scale = 0.4]{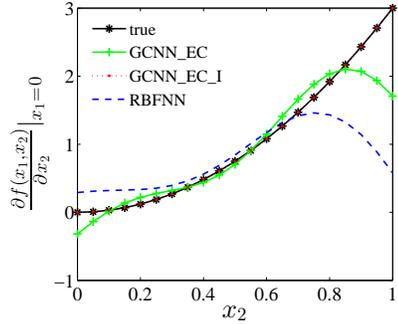}
\caption{Plots on the boundary $(x_{1}=0, x_{2})$ with the Neumann constraint.}
\label{PDE_neumann}
\end{figure}

Table \ref{tabel_Neumann} shows the performance in the boundary and the testing data with a Neumann boundary condition. A specific examination is made on the constraint boundary.
Fig. \ref{PDE_neumann} depicts the plots of three methods with the Neumann boundary condition. Obviously, GCNN\_EC\_I can satisfy the constraint exactly in the boundary because the Neumann constraint in Eq. (\ref{eq:objective of Nu}) is integrable for achieving an explicit expression. GCNN\_EC\_I is the best in solving the problem (\ref{eq:objective of Nu}). However, sometimes, an explicit expression may be unavailable or impossible so that GCNN\_EC is also a good choice. Note that a Neumann constraint is more difficult to be satisfied than a Dirichlet one.
GCNN\_EC presents a reasonable approximation except for the two ending ranges in the boundary.

\begin{table*}[htbp]
\centering
\scriptsize
\caption{Results for a PDE example with the Neumann boundary condition. ($N_{train}$ is the number of training data, $N_{test}$ is the number of testing data, $N_{RBF}$ is the number of RBF, $N_{pwc}$ is the number of point-wise constraints along the boundary. MSE means the average on the 100 groups of test data, $MSE\_{cstr}$ is the MSE on the constraints, $MSE\_{test}$ is the MSE on testing data.)}
\label{tabel_Neumann}
\begin{tabular}{|c|c|c|c|c|c|c|}
\hline
 Method &$N_{train}$& $N_{test}$ &$N_{RBF}$ & Key parameter & $MSE\_{cstr}$ &$MSE\_{test}$\\
\hline
RBFNN&121&321&$10$ & & $0.7081 $ & $0.0022 $  \\
\hline
GCNN\_EC& 121&321&$10$ & $\gamma=0.5$ &0.1693 &0.0167  \\
\hline
GCNN\_EC\_I& 121&321&$10$ & $\gamma=0.5$ &$\textbf{0}$ & $\textbf{0.0003}$  \\
\hline
\end{tabular}
\end{table*}

\section{Discussions of locality principle and coupling forms}
\label{sec:Discussions}

This section is an attempt to discuss locality principle from a viewpoint of constraint imposing in ANNs and 
to provide graphical interpretations about the differences between GIS and LIS. One typical question likes ``{\it how to discover Lagrange multiplier method to be GIS or LIS?}''. 
To answer this question, however, the interpretations are coupling-form dependent. 

One can show the original coupling form for the three methods in Table IV, but not for Lagrange multiplier method and GCNN-LP. The final prediction output $f(\textbf{x})$ contains two terms, where $f_{0}(\textbf{x})$ is a RBF output and $g_{s}(\textbf{x})$ is a superposition constraint. For the same methods, an alternative coupling form can be shown in Table V, where the alternative coupling term $G_{s}(\textbf{x})$ is different with $g_{s}(\textbf{x})$ in their expressions.
More specific forms of BVC-RBF and GCNN + Lagrange interpolation were discussed in \cite{hu2009a} and \cite{hong2009new}, respectively. The form of GCNN\_EC is equal to Eq. (\ref{eq: modified output function}).


\begin{table}
\scriptsize
\caption{Original Coupling Form ($f_0(\textbf{x})$ is a RBF ouput).}
\label{tabel_imposing}
\begin{tabular}{|c|c|}
  \hline
  Methods & Coupling of multiplication and superposition\\
   \hline
  BVC-RBF\cite{hong2009new} & $f(\textbf{x})= h(\textbf{x})f_{0}(\textbf{x})+g_s(\textbf{x})$\\
   \hline
  GCNN+Lagrange interpolation\cite{hu2009a} & $f(\textbf{x})= R_{1}(\textbf{x})f_{0}(\textbf{x})+g_s(\textbf{x})$\\
   \hline
  GCNN\_EC & $f(\textbf{x})=(1-\Psi(\textbf{x}))f_{0}(\textbf{x})+g_s(\textbf{x})$ \\
  \hline
\end{tabular}
\end{table}

\begin{table}
\begin{center} 
\scriptsize
\caption{Alternative Coupling Form by $f(\textbf{x})=f_0(\textbf{x})+G_{s}(\textbf{x})$.}
\label{tabel_superposition}
\begin{tabular}{|c|c|}
\hline
  Methods & Alternative coupling term for $G_{s}$\\
   \hline
  BVC-RBF\cite{hong2009new} & $h(\textbf{x})f_{0}(\textbf{x})+g(\textbf{x})-f_{0}(\textbf{x})$\\
   \hline
  GCNN+Lagrange interpolation\cite{hu2009a} & $R_{1}(\textbf{x}) f_{0}(\textbf{x})+R_{2}(\textbf{x})-f_{0}(\textbf{x})$ \\
   \hline
  GCNN\_EC & $\Psi(\textbf{x})(f_{C}(\textbf{x})-f_{0}(\textbf{x}))$ \\
  \hline
\end{tabular}
\end{center} 
\end{table}


\begin{figure*}[!ht]
 \centering
 \setlength{\abovecaptionskip}{0pt}
\setlength{\belowcaptionskip}{-9pt}
 \subfigure[BVC-RBF\cite{hong2009new}]{\label{subfig:curve_bvc-rbf1}
\includegraphics[scale = 0.30]{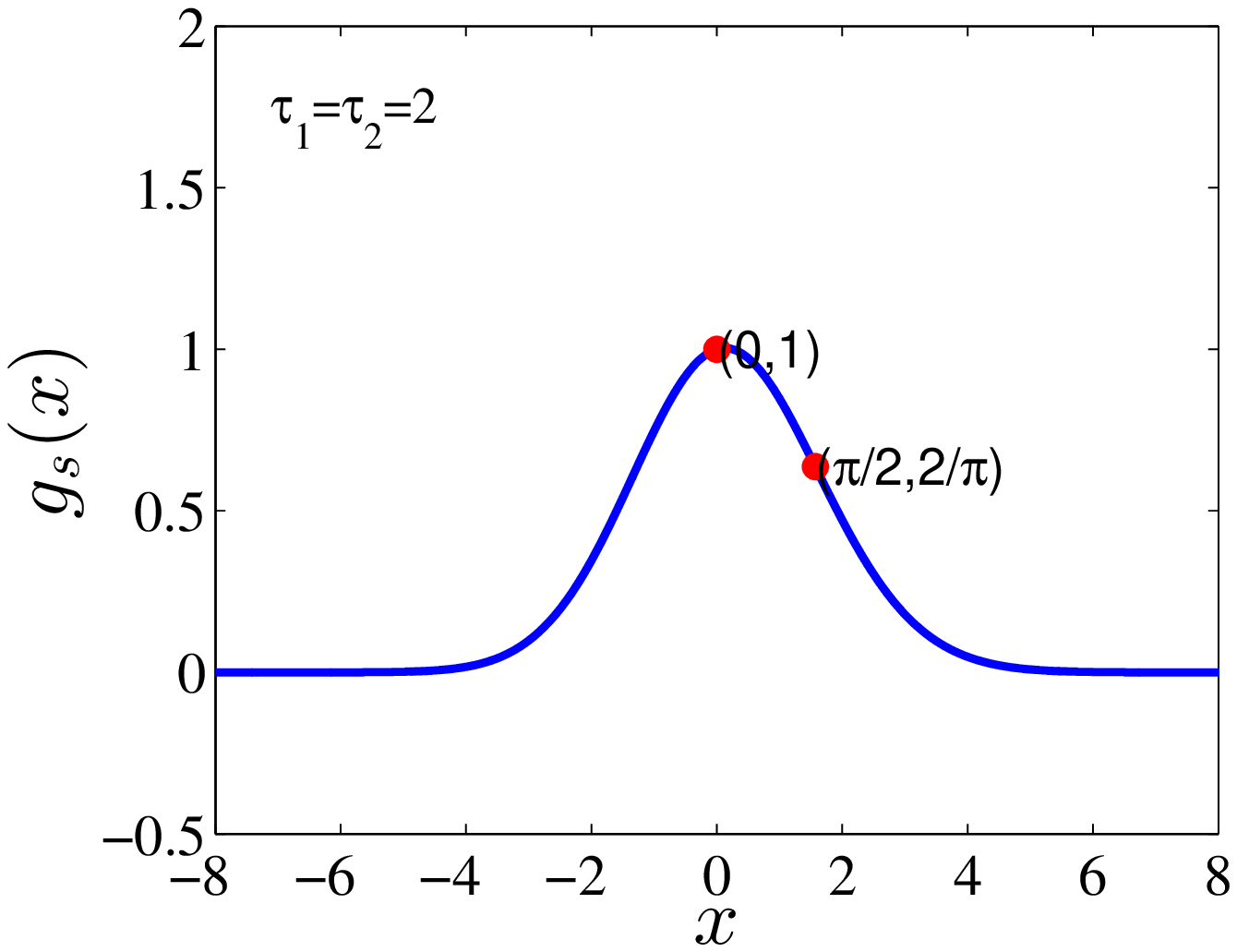}
 }
 \subfigure[GCNN+Lagrange interpolation\cite{hu2009a}]{\label{subfig:curve_gcnn1}
\includegraphics[scale = 0.30]{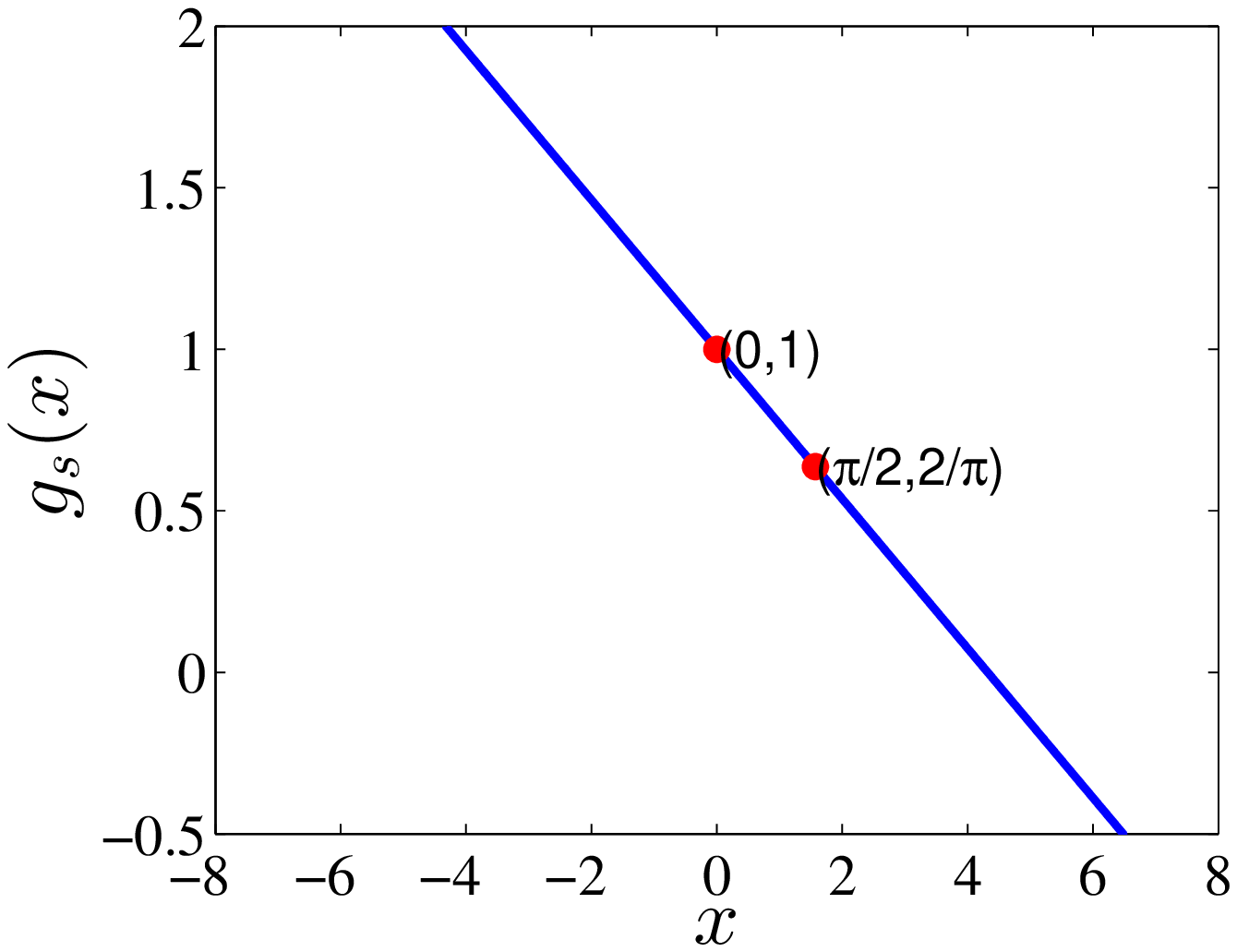}
 }
  \subfigure[GCNN\_EC]{\label{subfig:curve_gcnnec1}
\includegraphics[scale = 0.30]{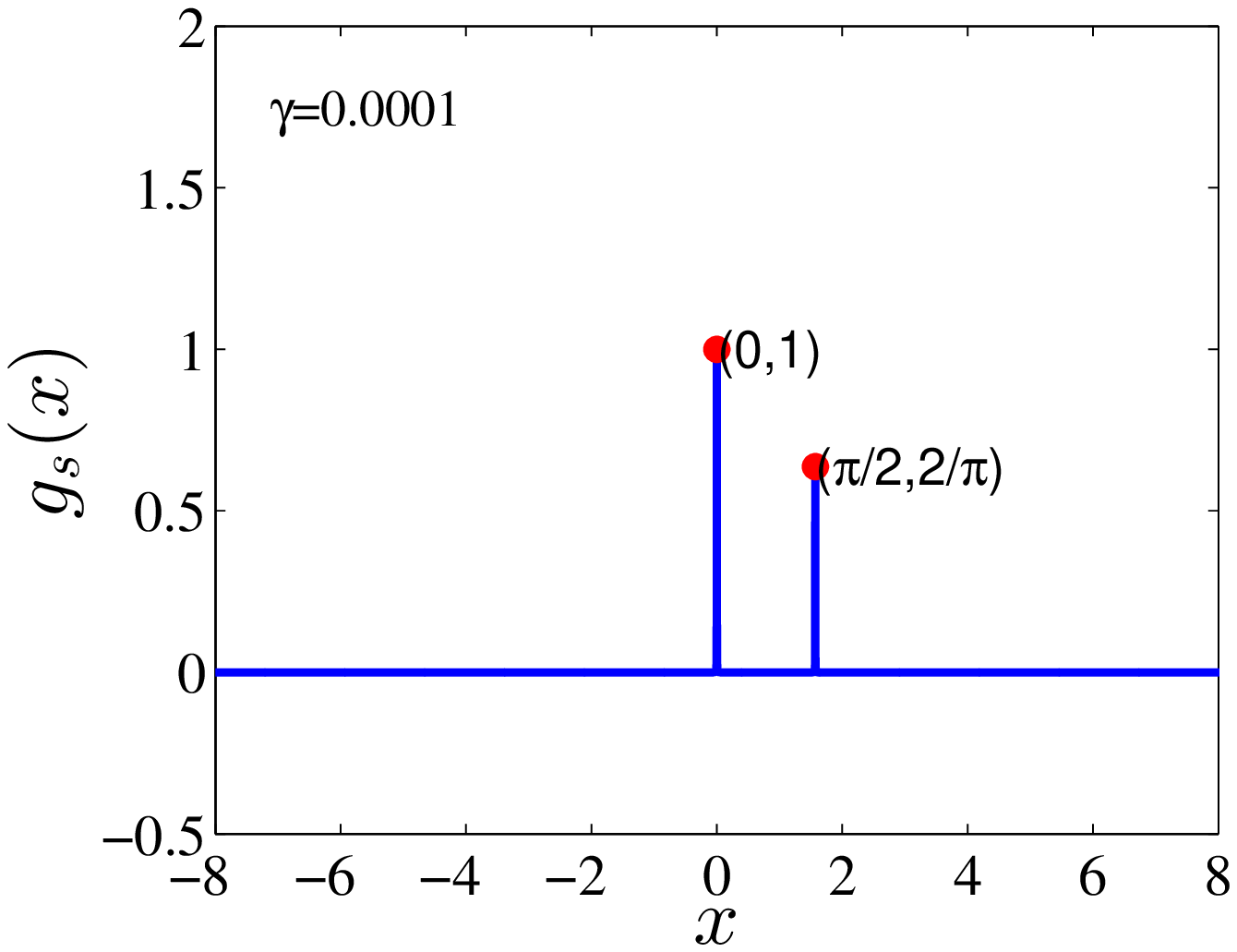}
 }
  \centering
 \caption{$g_{s}({x})$ plots of BVC-RBF, GCNN+Lagrange interpolation and GCNN\_EC in the original coupling form for a {\it Sinc} function in which two constraints
 are located at $x=0$ and $x=\pi/2$, respectively.}
 \label{fig:imposing}
 \end{figure*}

For a better understanding about differences among the given three methods, 
we set the $Sinc$ function as an example, in which
two interpolation point constraints are enforced but without additive noise. 
Fig. 4 shows the original coupling function $g_{s}({x})$, and
Fig. 5 shows both RBF output  $f_{0}({x})$ and alternative coupling function $G_{s}({x})$ together. We keep parameters $\tau_{1}=\tau_{2}=2$ for BVC-RBF 
for reason of good performance on the data. When $\tau_{1}=\tau_{2}<1$, the performance becomes poor. Within either of the coupling forms,
GCNN\_EC presents the best in terms of locality from $g_{s}({x})$ or $G_{s}({x})$. The plots confirm that the locality interpretations are coupling-form dependent.

\begin{figure}[!htbp]
 \setlength{\abovecaptionskip}{0pt}
\setlength{\belowcaptionskip}{-9pt}
  \subfigure[$f_{0}(x)$ of BVC-RBF\cite{hong2009new}]{\label{subfig:f0_bvc-rbf}
\includegraphics[width=1.61in]{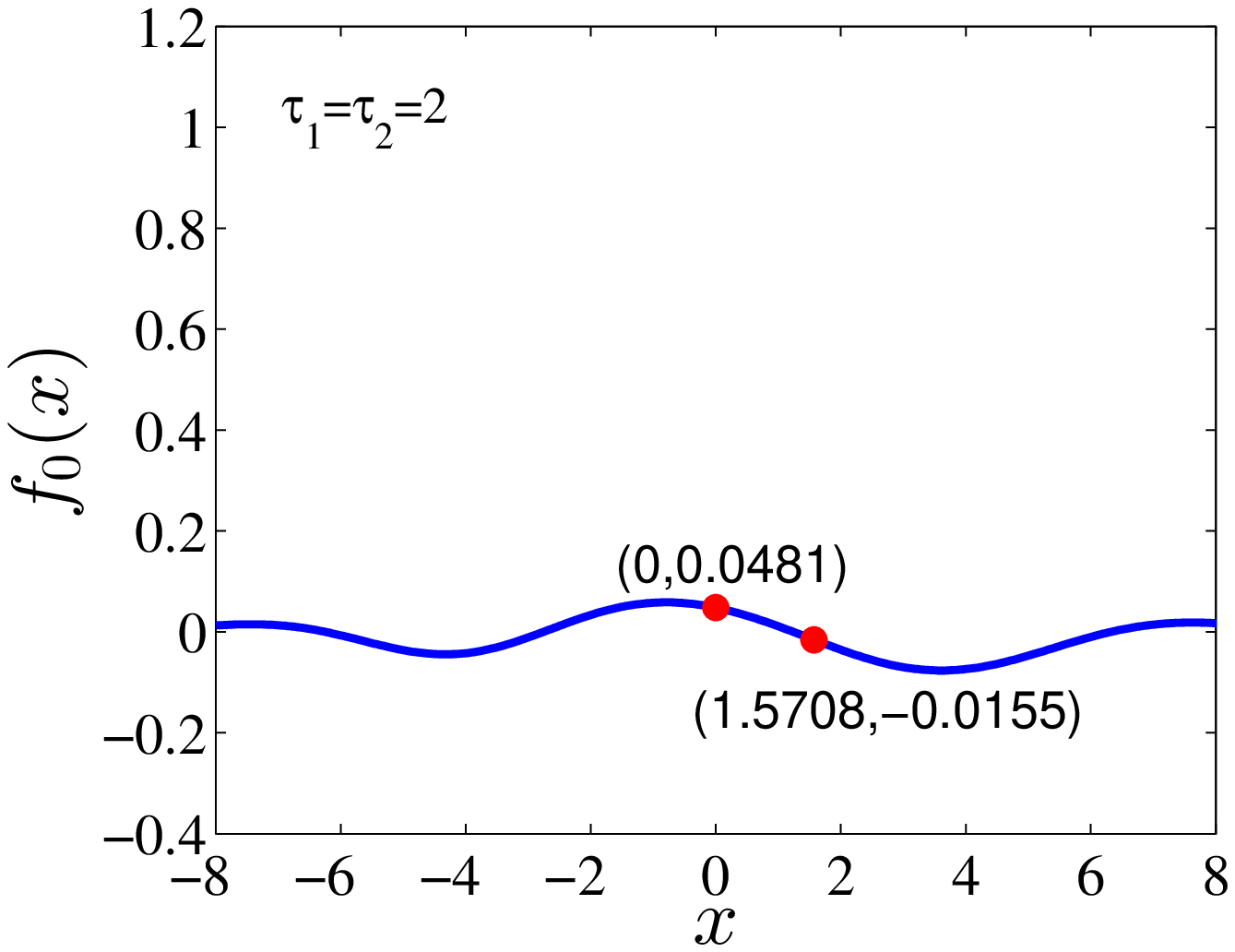}
 }
 \subfigure[$G_{s}(x)$ of BVC-RBF\cite{hong2009new}]{\label{subfig:Gs_bvc-rbf}
\includegraphics[width=1.61in]{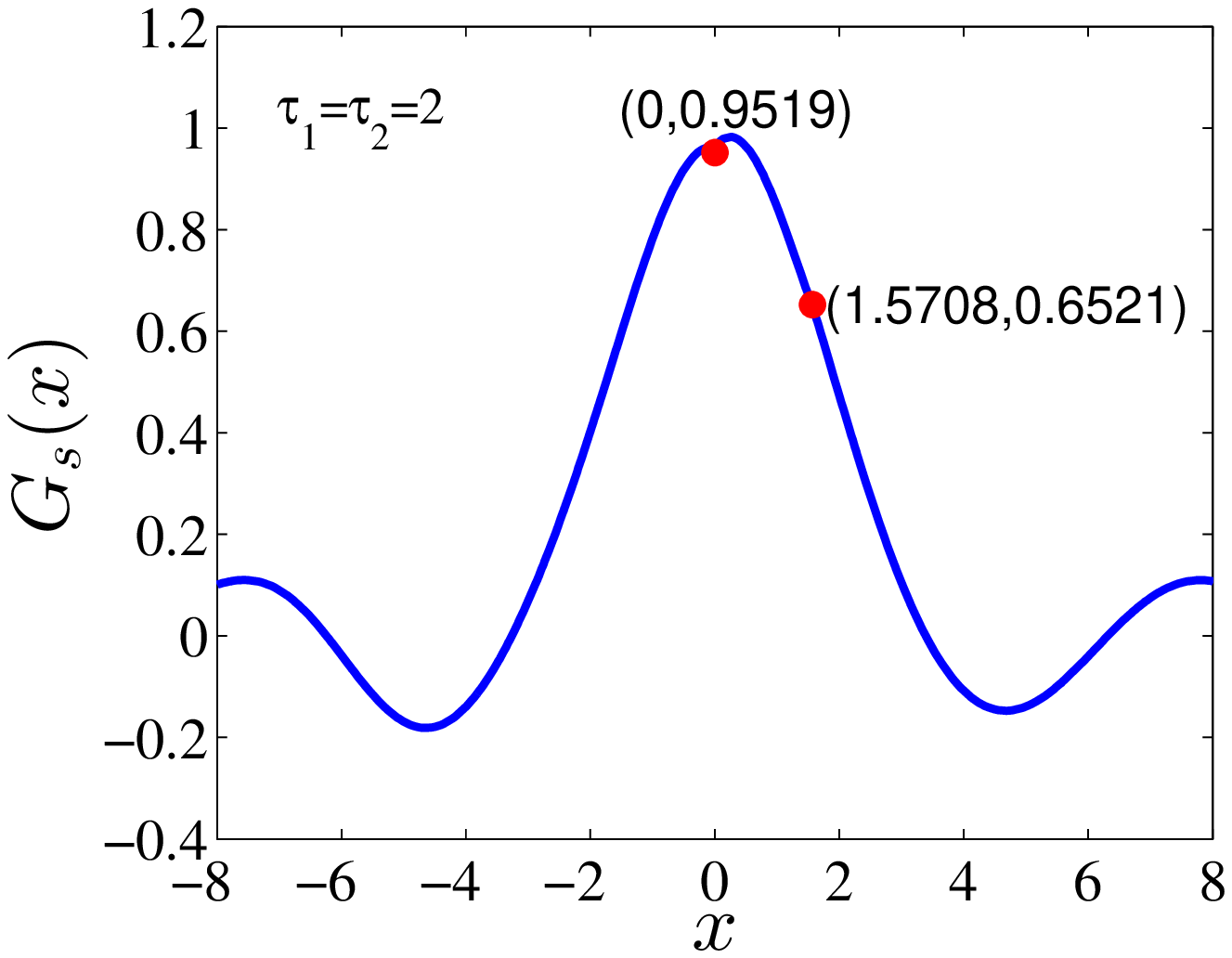}
 }\\
  \subfigure[$f_{0}(x)$ of GCNN + Lagrange interpolation\cite{hu2009a}]{\label{subfig:f0_gcnn}
\includegraphics[width=1.61in]{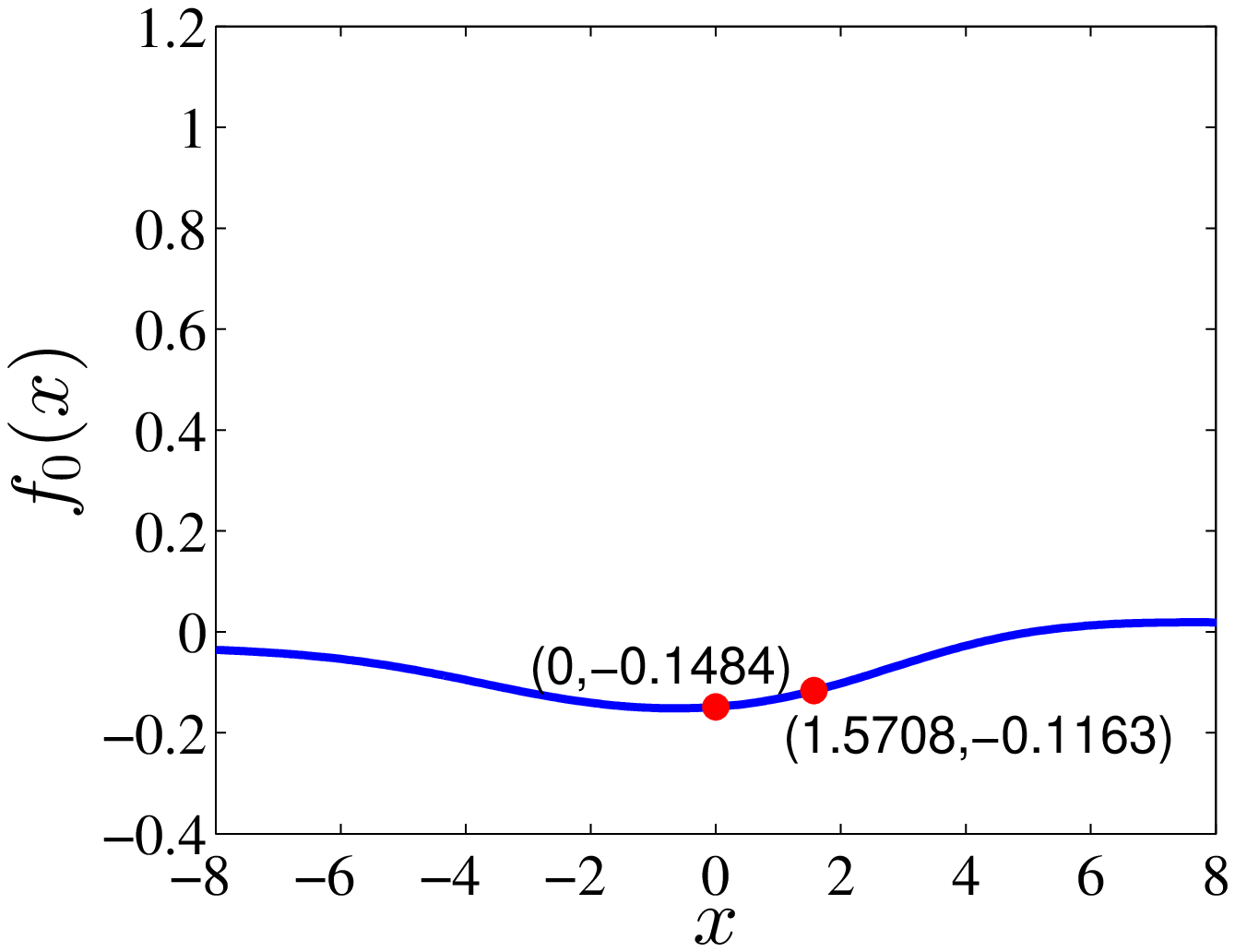}
 }
 \subfigure[$G_{s}(x)$ of GCNN + Lagrange interpolation\cite{hu2009a}]{\label{subfig:Gs_gcnn}
\includegraphics[width=1.61in]{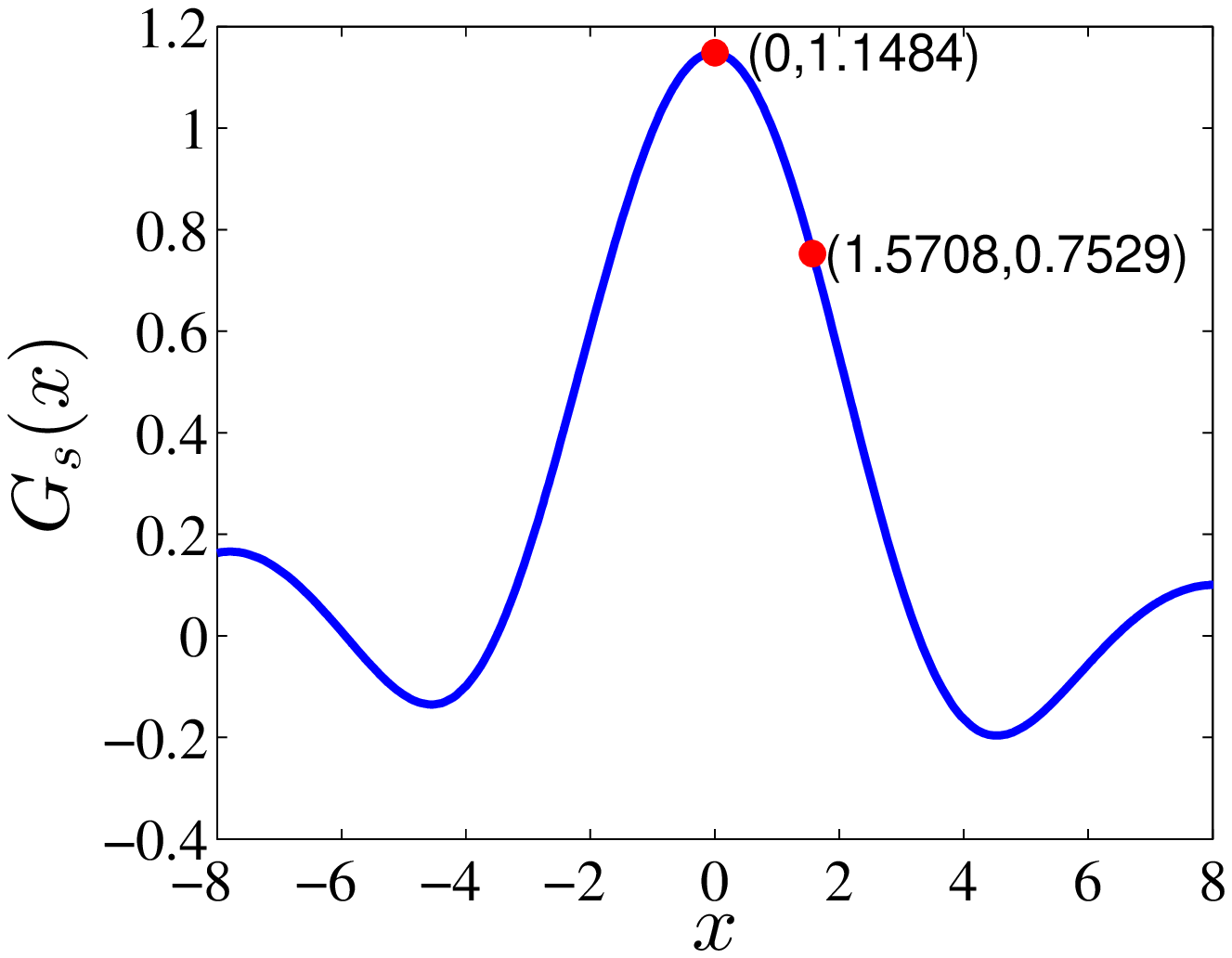}
 }\\
  \subfigure[$f_{0}(x)$ of GCNN\_EC]{\label{subfig:f0_gcnnec}
\includegraphics[width=1.61in]{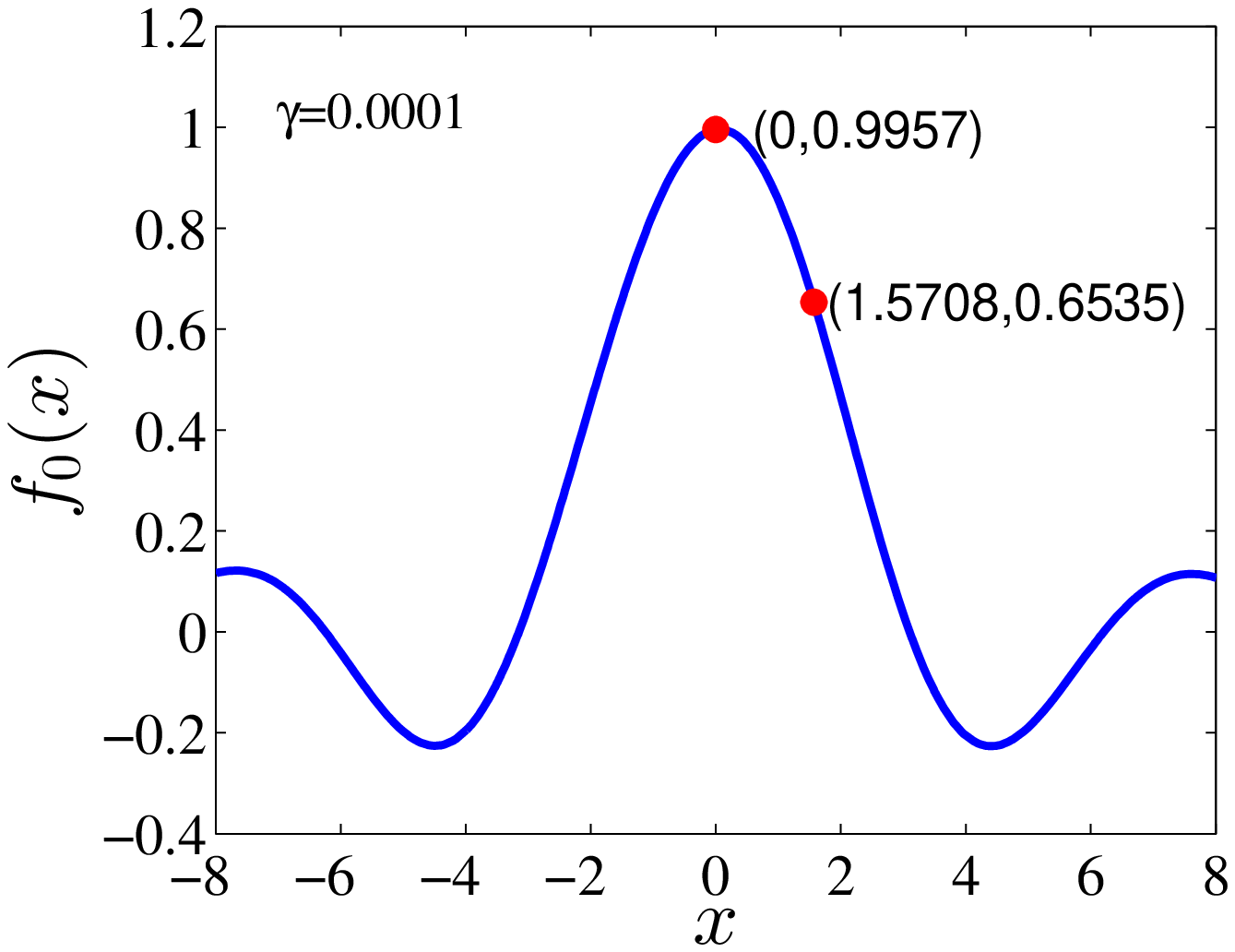}
 }
 \subfigure[$G_{s}(x)$ of GCNN\_EC]{\label{subfig:Gs_gcnnec}
\includegraphics[width=1.61in]{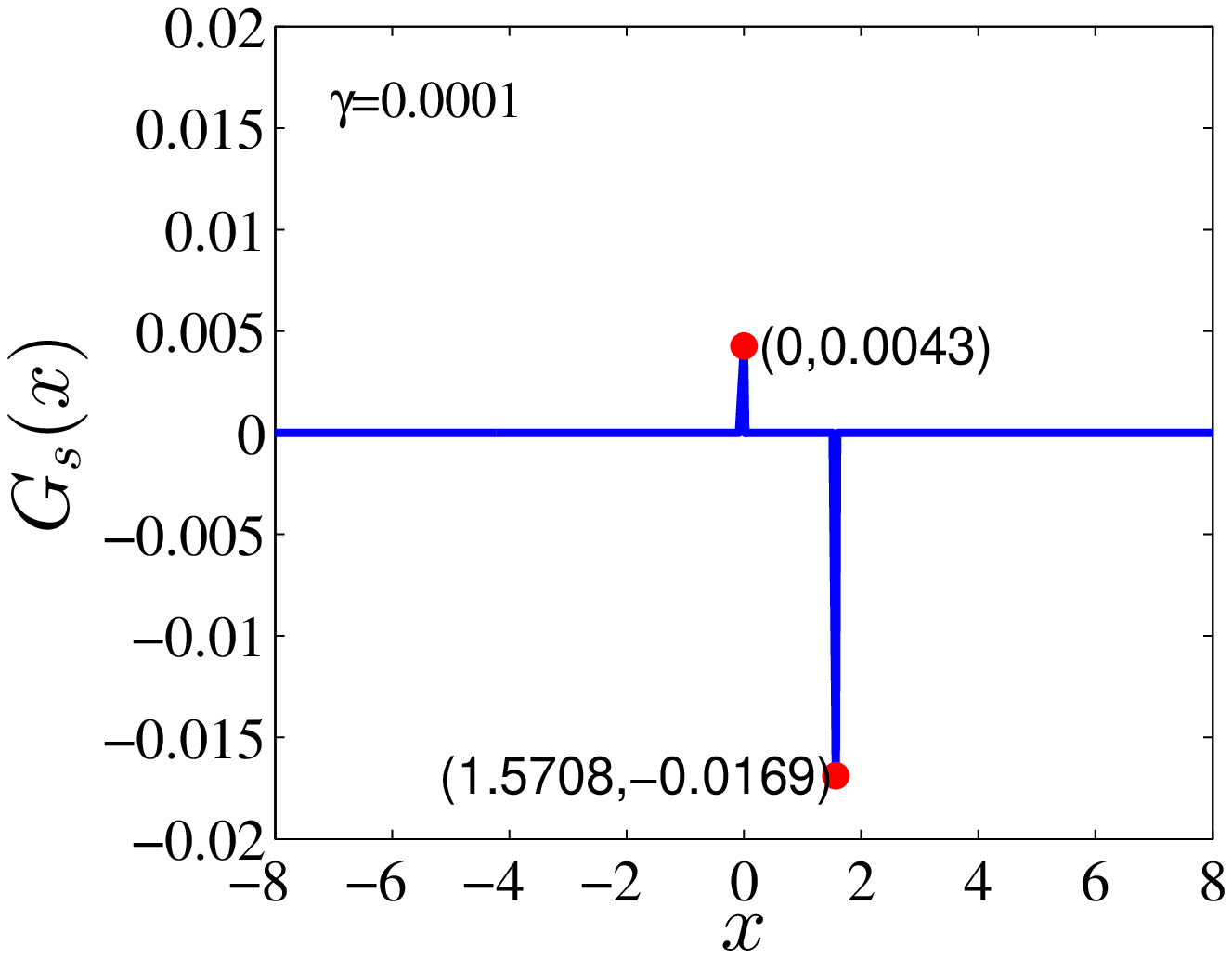}
 }
 \caption{$f_{0}(x)$ and $G_{s}(x)$ plots of BVC-RBF, GCNN + Lagrange interpolation and GCNN\_EC in an alternative coupling form for a {\it Sinc} function in which two constraints
 are located at $x=0$ and $x=\pi/2$, respectively.}
 \label{fig:modification}
 \end{figure}

However, one is unable to derive such explicit forms, either $g_{s}(\textbf{x})$ or $G_{s}(\textbf{x})$, for Lagrange multiplier method and GCNN-LP. 
In order to reach an overall comparison about them, we propose a generic coupling form in the following expression:
 \begin{equation}
 \label{eq:superposition}
 \begin{split}
 f(\textbf{x})&= f_{wc}(\textbf{x})+ f_{m}(\textbf{x}),
 \end{split}
 \end{equation}
where $f_{m}(\textbf{x})$ is the modification output over the RBF output $f_{wc}(\textbf{x})$ without constraints. 
One can imagine that the given constraints work as a modification function $f_{m}(\textbf{x})$
and impose it additively on the original RBF output $f_{wc}(\textbf{x})$ to form the final prediction output $f(\textbf{x})$.
All constraint methods can be examined by Eq. (\ref{eq:superposition}).  
However, this examination is basically a numerical one and requires an extra calculation of $f_{wc}(\textbf{x})$. 
Fig. 6 shows the plots of $f_m$ from RBFNN+Lagrange multiplier and GCNN\_EC models. One can observe their significant differences in locality behaviors.

\begin{figure}[!htb]
 \setlength{\abovecaptionskip}{0pt}
\setlength{\belowcaptionskip}{-9pt}
\subfigure[RBFNN+Lagrange multiplier]{\label{subfig:fm_Lagrange}
\includegraphics[width=1.63in]{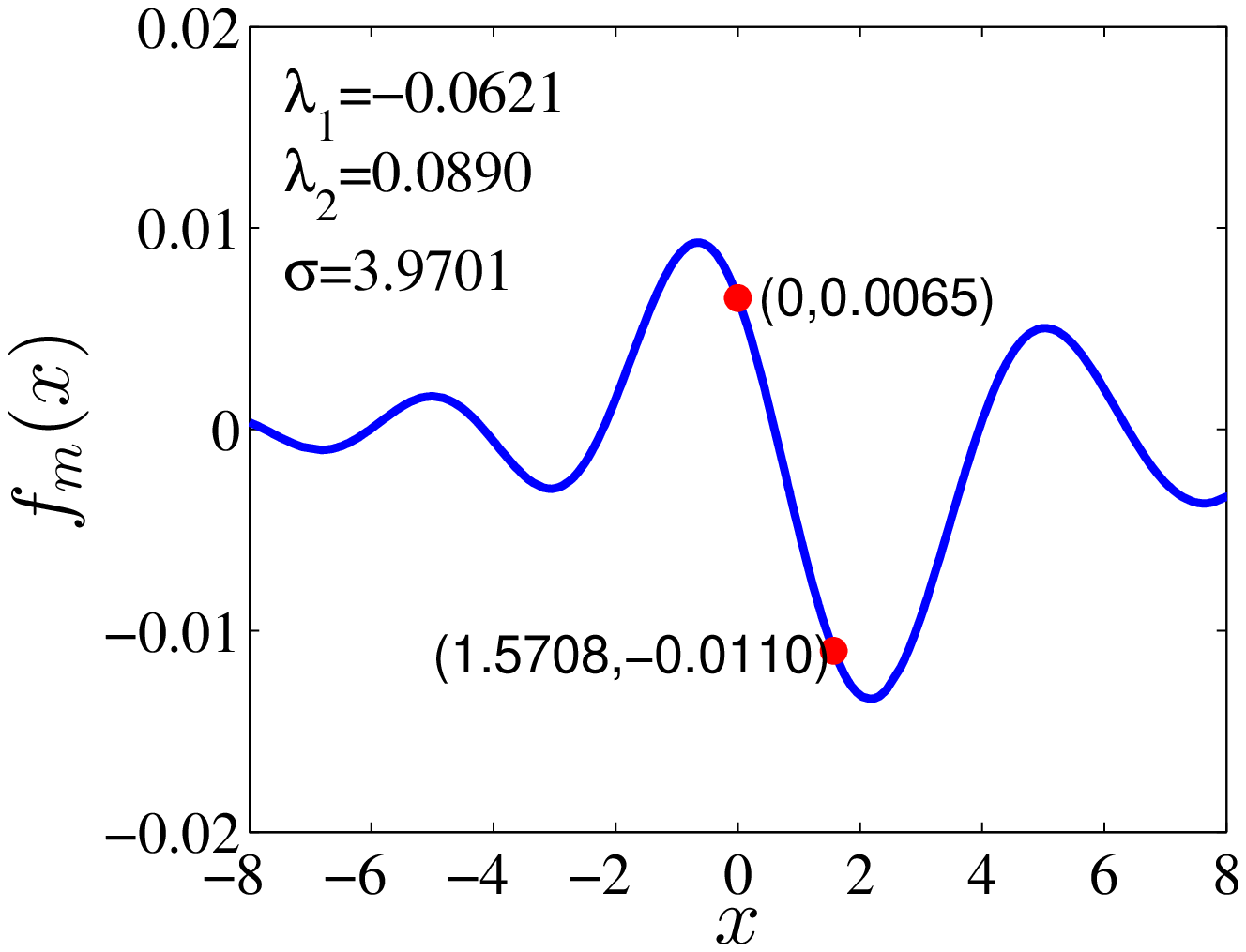}
 }
  \subfigure[GCNN\_EC]{\label{subfig:fm_GCNNEC}
\includegraphics[width=1.63in]{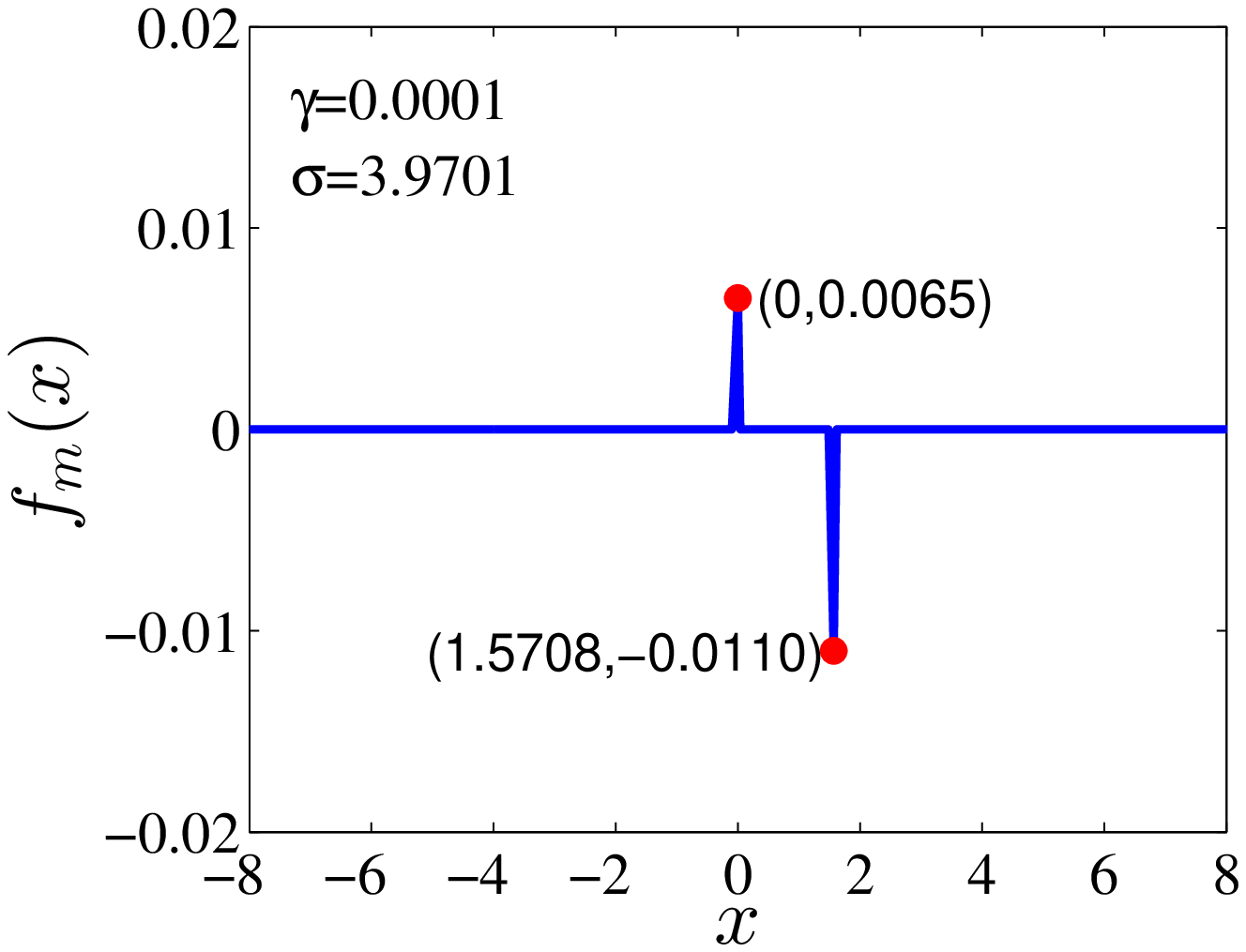}
 }
 \caption{$f_{m}$ plots of RBFNN+Lagrange multiplier and GCNN\_EC in the generic coupling form for a {\it Sinc} function in which two constraints
 are located at $x=0$ and $x=\pi/2$, respectively.}
 \label{fig:f_m}
 \end{figure}

In this work, $f(\textbf{x})$ and $f_{wc}(\textbf{x})$ represent
two RBF neural networks with and without constraints, respectively. Because brain memory is attributed to the changes in {\it synaptic strength} or {\it connectivity} \cite{destexhe2004plasticity}, we propose the following steps in designs of the two networks. 
First, the same number of neurons is applied so that they share the same connectivity in terms of neurons (but not in terms of constrains).  
Second, the same preset values on the parameters $\boldsymbol{\mu}_j$ and ${\sigma}_j$ are given respectively to the two networks. 
Step 3, the weight parameters $w_j$ of GCNN\_EC are gained from solving a linear problem which guarantees a unique solution. 
Lagrange multiplier method will take the weights obtained from $f_{wc}(\textbf{x})$ as an initial condition for updating $w_j$ in $f(\textbf{x})$. 
The updating operation is to emulate a brain memory change.  
The above steps will enable us to examine the changes from synaptic strengths (or weight parameters) between the two networks. 

When Figs. 4 to 6 provide a locality interpretation from a ``{\it signal function''} sense, another interpretation is explored from the plots of ``{\it weight changes''} between $f_{wc}(\textbf{x})$
and $f_{}(\textbf{x})$. Because the two networks have the same number of neurons or weight parameters, we denote $\Delta W$ to be their weight changes. 
Normalized weight changes will be achieved for $\Delta W/|\Delta W|_{max}$, where $|\Delta W|_{max}$ is a normalization scalar. 
We still take the $Sinc$ function for an example. Comparisons are made again between RBFNN + Lagrange multiplier method and GCNN\_EC. 
Fig. 6 shows the plots of normalized weight changes of RBFNN + Lagrange multiplier and GCNN\_EC. 
Numerical tests indicate that behavior of locality property in the plots is dependent to some parameters of networks. 
For reaching meaningful plots, we set $N_{RBF}=500$, and $N_{train}=1000$. The center parameters ${\mu}_j$ are generated uniformly along $x$ variable at the intervals [−10, 10] so that the center interval is about 0.04. 
The constant $\sigma(=\sigma_j)$ is given with values of 0.05, 0.10 and 0.15, respectively. When $\sigma$ is decreased (say, equal to the center interval), the performance becomes poor for both RBFNN + Lagrange multiplier and GCNN\_EC. 
 
From Fig. 6 one can observe that, when $\sigma=0.05$, both RBFNN + Lagrange multiplier and GCNN\_EC show the locality property on the constraint locations. When $\sigma=0.10$ or $0.15$, RBFNN + Lagrange multiplier loses the locality property, but GCNN\_EC is in a less degree. Numerical tests imply that GCNN\_EC holds a locality property better than RBFNN + Lagrange multiplier. 

From the discussions so far, we can ensure the differences between GIS and LIS, but still cannot answer the question given in this section. It is an open problem requiring both theoretical and numerical findings.  

\begin{figure}[!htbp]
 \setlength{\abovecaptionskip}{0pt}
\setlength{\belowcaptionskip}{-9pt}
  \subfigure[RBFNN + Lagrange multiplier: $\sigma=0.05$ ]{\label{subfig:delta_W_Lagrange_0.05}
\includegraphics[width=1.63in]{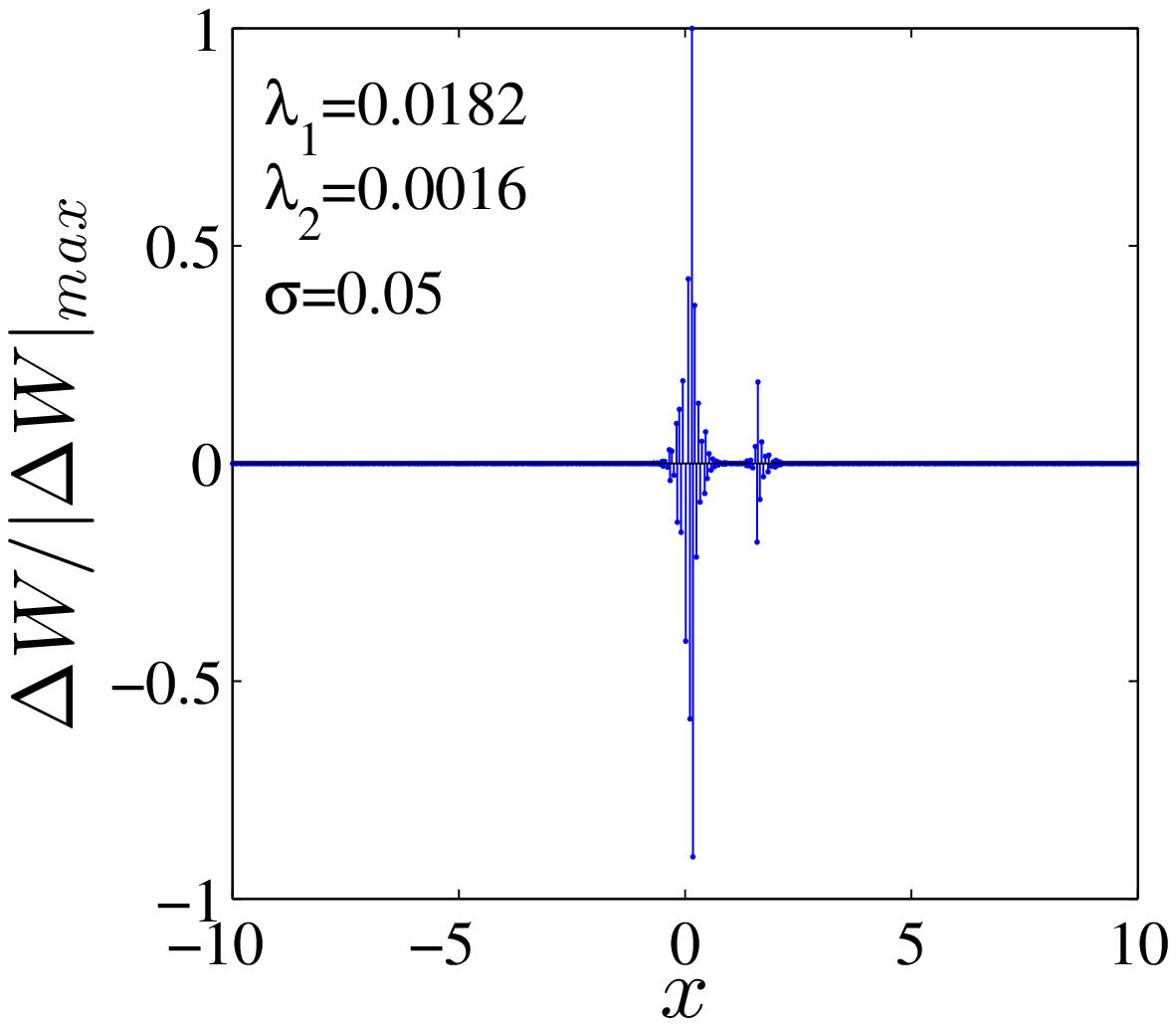}
 }
 \subfigure[GCNN\_EC: $\sigma=0.05$]{\label{subfig:delta_W_GCNNEC_0.05}
\includegraphics[width=1.63in]{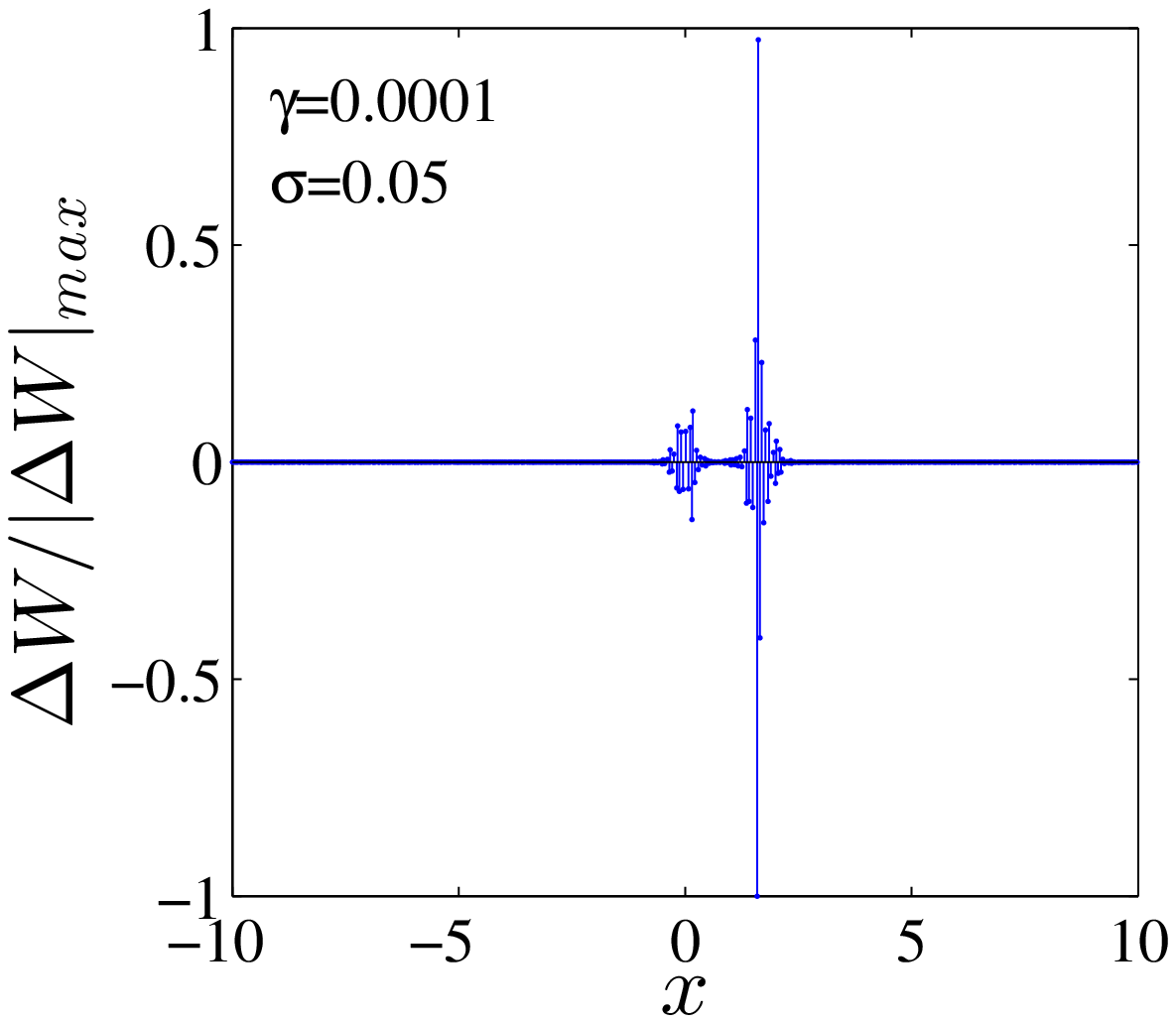}
 }\\
 \subfigure[RBFNN + Lagrange multiplier: $\sigma=0.10$ ]{\label{subfig:delta_W_Lagrange_0.10}
\includegraphics[width=1.63in]{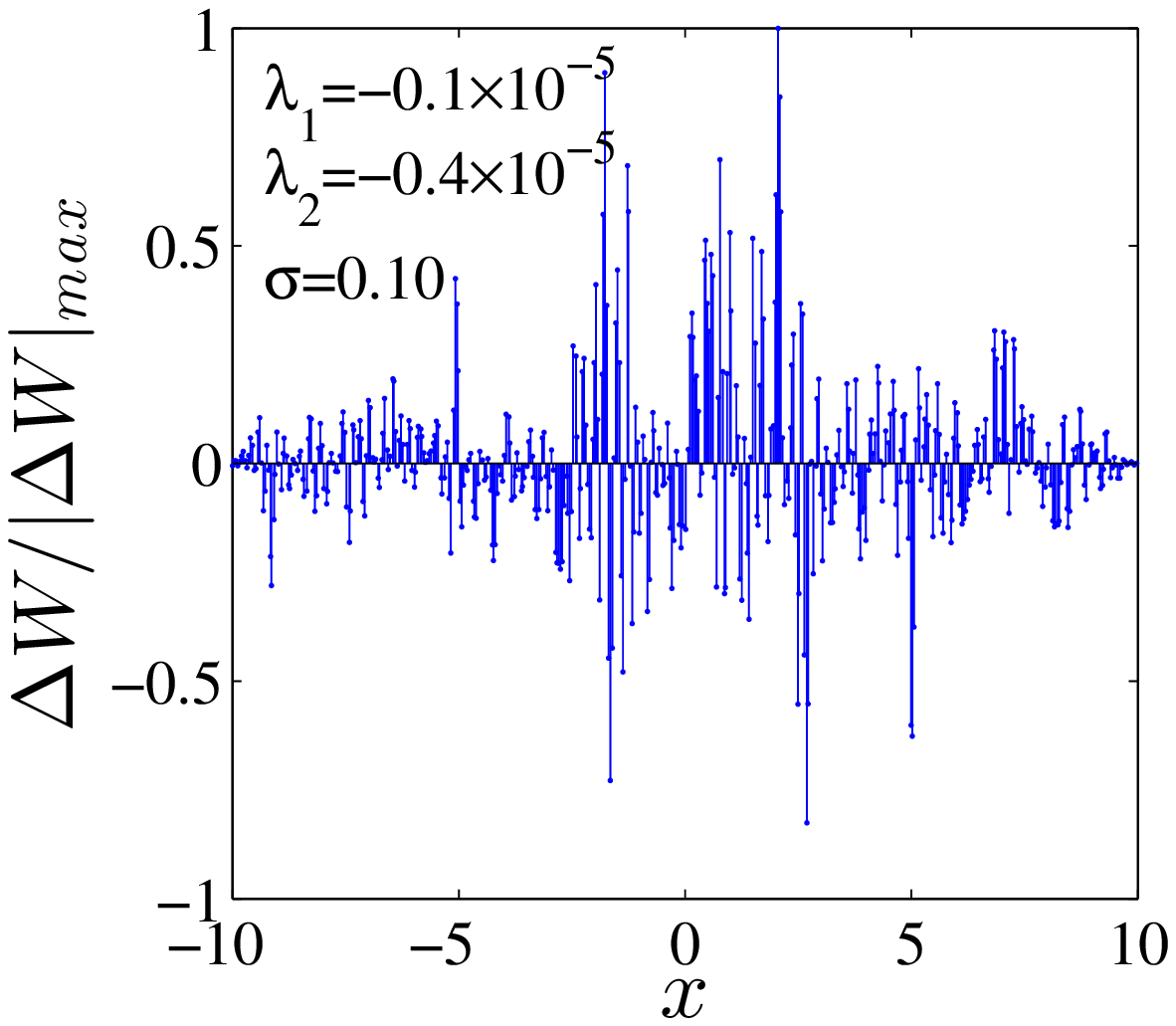}
 }
   \subfigure[GCNN\_EC: $\sigma=0.10$]{\label{subfig:delta_W_GCNNEC_0.10}
\includegraphics[width=1.63in]{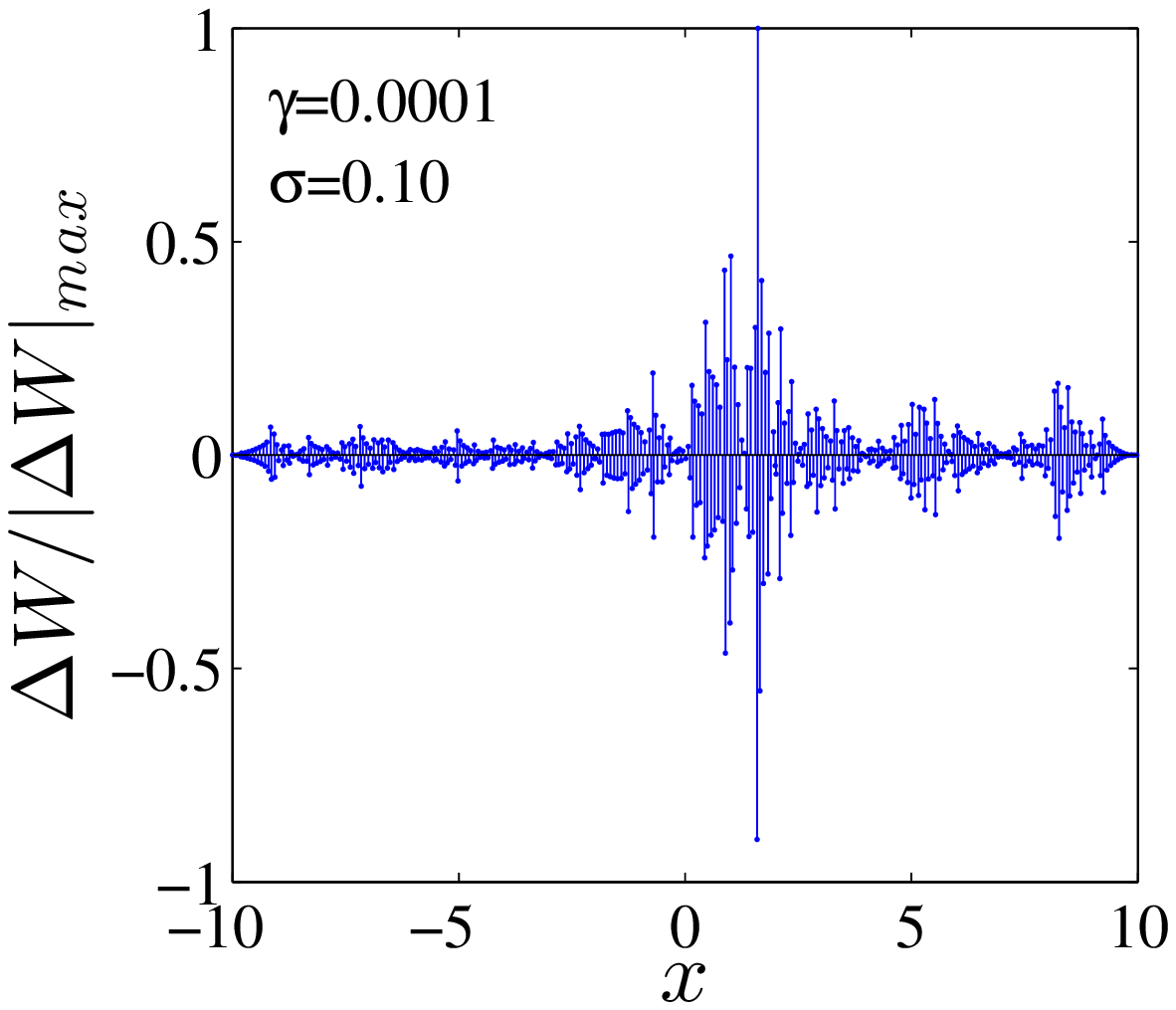}
 }\\
 \subfigure[RBFNN + Lagrange multiplier: $\sigma=0.15$ ]{\label{subfig:delta_W_Lagrange_0.15}
\includegraphics[width=1.63in]{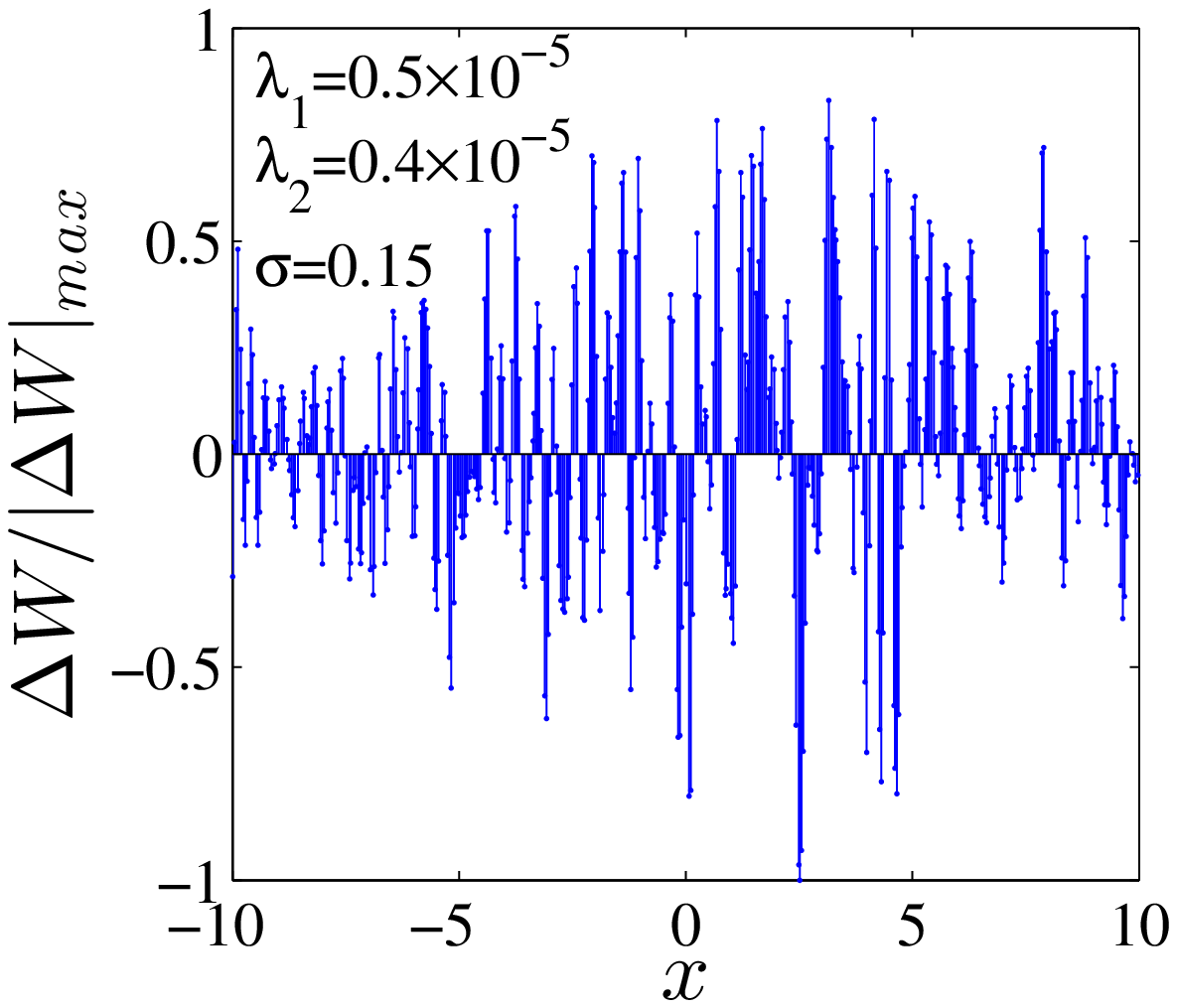}
 }
   \subfigure[GCNN\_EC: $\sigma=0.15$]{\label{subfig:delta_W_GCNNEC_0.15}
\includegraphics[width=1.63in]{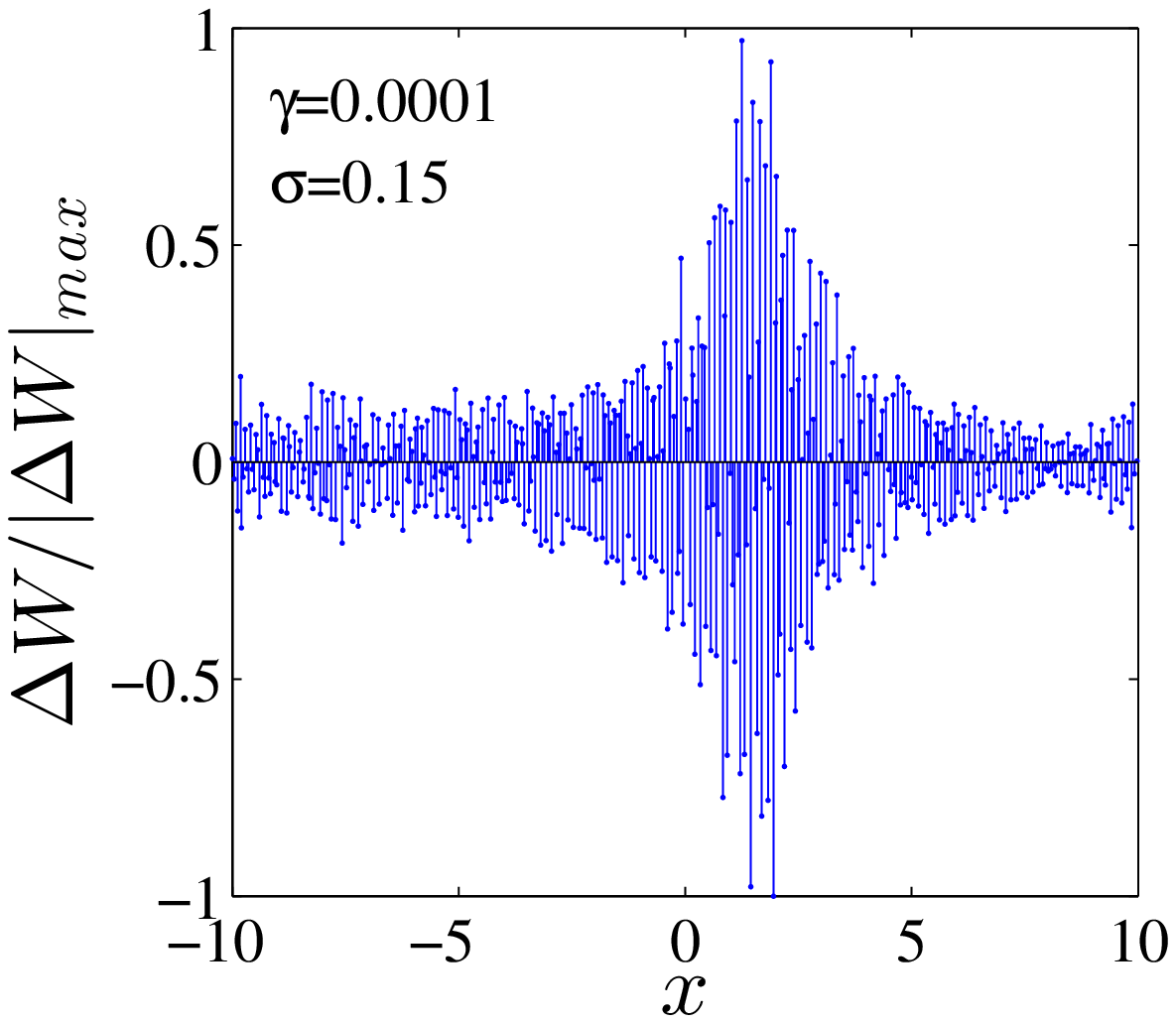}
}
 \caption{Normalized weight changes of RBFNN + Lagrange multiplier and GCNN\_EC for a {\it Sinc} function in which two constraints
 are located at $x=0$ and $x=\pi/2$, respectively.}
 \label{fig:delta_W}
 \end{figure}
 
\section{Final remarks}
\label{sec:remarks}

In this work, we study on the constraint imposing scheme of the GCNN models. We first discuss the geometric differences between the conventional optimization problems and machine learning problems. Based on the discussions, a new method within LIS is proposed for the GCNN models. GCNN\_EC transfers
equality constraint problems into unconstrained ones and solves them by a linear approach, so that convexity of constraints is no more an issue. 
The present method is able to process interpolation function constraints that cover the constraint types in BVPs.
Numerical study is made by including the constraints in the forms of Dirichlet and Neumann for the BVPs.
GCNN\_EC achieves an exact satisfaction of the equality constraints, with either Dirichlet or Neumann types, when they are expressed by an explicit form about $f$. The approximations are obtained if a Neumann constraint is not integrable for an explicit form about $f$. 

A numerical comparison is made for the methods within GIS and LIS. Graphical interpretations are given to show that the locality principle in the brain study has a wider meaning in ANNs. In apart from local properties in CNN \cite{le1990handwritten} and RBF \cite{schwenker2001three}, coupling forms between knowledge and data can be another locality source for studies. We believe that the locality principle is one of key steps for ANNs to realize a brain-inspired machine. 
The present work indicates a new direction for advancing ANN technique. When Lagrange multiplier is
a standard method in machine learning, we show that LIS can be an alternative solution and can performance better in the given problems.  We need to explore LIS and GIS together and try to understand under which conditions LIS or GIS should be selected.


\section*{acknowledgment}

Thanks to Dr. Yajun Qu, Guibiao Xu and Yanbo Fan for the helpful discussions.
The open-source code, GCNN-LP, developed by Yajun Qu is used (http://www.openpr.org.cn/).
This work is supported in part by NSFC No. 61273196 and 61573348.



%

%
%
\bibliographystyle{IEEEtran}
\bibliography{reference}

\begin{thebibliography}{10}
\providecommand{\url}[1]{#1}
\csname url@samestyle\endcsname
\providecommand{\newblock}{\relax}
\providecommand{\bibinfo}[2]{#2}
\providecommand{\BIBentrySTDinterwordspacing}{\spaceskip=0pt\relax}
\providecommand{\BIBentryALTinterwordstretchfactor}{4}
\providecommand{\BIBentryALTinterwordspacing}{\spaceskip=\fontdimen2\font plus
\BIBentryALTinterwordstretchfactor\fontdimen3\font minus
  \fontdimen4\font\relax}
\providecommand{\BIBforeignlanguage}[2]{{%
\expandafter\ifx\csname l@#1\endcsname\relax
\typeout{** WARNING: IEEEtran.bst: No hyphenation pattern has been}%
\typeout{** loaded for the language `#1'. Using the pattern for}%
\typeout{** the default language instead.}%
\else
\language=\csname l@#1\endcsname
\fi
#2}}
\providecommand{\BIBdecl}{\relax}
\BIBdecl

\bibitem{hu2009a}
B.-G. Hu, H.~B. Qu, Y.~Wang, and S.~H. Yang, ``A generalized-constraint neural
  network model: Associating partially known relationships for nonlinear
  regressions,'' \emph{Information Sciences}, vol. 179, no.~12, pp. 1929--1943,
  2009.

\bibitem{cao2015generalized}
L.~L. Cao and B.-G. Hu, ``Generalized constraint neural network regression
  model subject to equality function constraints,'' in \emph{Proc. of
  International Joint Conference on Neural Networks (IJCNN)}, 2015, pp. 1--8.

\bibitem{lecun2015deep}
Y.~LeCun, Y.~Bengio, and G.~Hinton, ``Deep learning,'' \emph{Nature}, vol. 521,
  no. 7553, pp. 436--444, 2015.

\bibitem{Deng}
L.~Deng and D.~Yu, ``Deep learning: Methods and applications,''
  \emph{Foundations and Trends in Signal Processing}, vol.~7, no. 3-4, pp.
  197--387, 2014.

\bibitem{Schmidhuber}
J.~Schmidhuber, ``Deep learning in neural networks: An overview,'' \emph{Neural
  Networks}, vol.~61, pp. 85--117, 2015.

\bibitem{todorovski2006integrating}
L.~Todorovski and S.~D{\v{z}}eroski, ``Integrating knowledge-driven and
  data-driven approaches to modeling,'' \emph{Ecological Modelling}, vol. 194,
  no.~1, pp. 3--13, 2006.

\bibitem{olden2002illuminating}
J.~D. Olden and D.~A. Jackson, ``Illuminating the {``black box''}: a
  randomization approach for understanding variable contributions in artificial
  neural networks,'' \emph{Ecological Modelling}, vol. 154, no.~1, pp.
  135--150, 2002.

\bibitem{yang2008structural}
S.~H. Yang, B.-G. Hu, and P.~H. Courn{\`e}de, ``Structural identifiability of
  generalized constraint neural network models for nonlinear regression,''
  \emph{Neurocomputing}, vol.~72, no.~1, pp. 392--400, 2008.

\bibitem{qu2011generalized}
Y.-J. Qu and B.-G. Hu, ``Generalized constraint neural network regression model
  subject to linear priors,'' \emph{IEEE Transactions on Neural Networks},
  vol.~22, no.~12, pp. 2447--2459, 2011.

\bibitem{ran2014determining}
Z.-Y. Ran and B.-G. Hu, ``Determining structural identifiability of parameter
  learning machines,'' \emph{Neurocomputing}, vol. 127, pp. 88--97, 2014.

\bibitem{fan2015a}
X.-R. Fan, M.-Z. Kang, E.~Heuvelink, P.~de~Reffye, and B.-G. Hu, ``A
  knowledge-and-data-driven modeling approach for simulating plant growth: A
  case study on tomato growth,'' \emph{Ecological Modelling}, vol. 312, pp.
  363--373, 2015.

\bibitem{psichogios1992hybrid}
D.~C. Psichogios and L.~H. Ungar, ``A hybrid neural network-first principles
  approach to process modeling,'' \emph{AIChE Journal}, vol.~38, no.~10, pp.
  1499--1511, 1992.

\bibitem{thompson1994modeling}
M.~L. Thompson and M.~A. Kramer, ``Modeling chemical processes using prior
  knowledge and neural networks,'' \emph{AIChE Journal}, vol.~40, no.~8, pp.
  1328--1340, 1994.

\bibitem{zadeh1986outline}
L.~A. Zadeh, ``Outline of a computational approach to meaning and knowledge
  representation based on the concept of a generalized assignment statement,''
  in \emph{Proc. of the International Seminar on Artificial Intelligence and
  Man-Machine Systems}.\hskip 1em plus 0.5em minus 0.4em\relax Springer, 1986,
  pp. 198--211.

\bibitem{zadeh1996fuzzy}
------, ``Fuzzy logic = computing with words,'' \emph{IEEE Transactions on
  Fuzzy Systems}, vol.~4, no.~2, pp. 103--111, 1996.

\bibitem{denning2005locality}
P.~J. Denning, ``The locality principle,'' \emph{Communications of the ACM},
  vol.~48, no.~7, pp. 19--24, Jul. 2005.

\bibitem{boyd2004convex}
S.~Boyd and L.~Vandenberghe, \emph{Convex Optimization}.\hskip 1em plus 0.5em
  minus 0.4em\relax Cambridge University Press, 2004.

\bibitem{bishop2006pattern}
C.~M. Bishop, \emph{Pattern Recognition and Machine Learning}.\hskip 1em plus
  0.5em minus 0.4em\relax Springer, 2006.

\bibitem{lagaris1998artificial}
I.~E. Lagaris, A.~Likas, and D.~I. Fotiadis, ``Artificial neural networks for
  solving ordinary and partial differential equations,'' \emph{IEEE
  Transactions on Neural Networks}, vol.~9, no.~5, pp. 987--1000, 1998.

\bibitem{hong2009new}
X.~Hong and S.~Chen, ``A new {RBF} neural network with boundary value
  constraints,'' \emph{IEEE Transactions on Systems, Man, and Cybernetics, Part
  B: Cybernetics}, vol.~39, no.~1, pp. 298--303, 2009.

\bibitem{mcfall2009artificial}
K.~S. McFall and J.~R. Mahan, ``Artificial neural network method for solution
  of boundary value problems with exact satisfaction of arbitrary boundary
  conditions,'' \emph{IEEE Transactions on Neural Networks}, vol.~20, no.~8,
  pp. 1221--1233, 2009.

\bibitem{lauer2008incorporating}
F.~Lauer and G.~Bloch, ``Incorporating prior knowledge in support vector
  regression,'' \emph{Machine Learning}, vol.~70, no.~1, pp. 89--118, 2008.

\bibitem{schwenker2001three}
F.~Schwenker, H.~A. Kestler, and G.~Palm, ``Three learning phases for
  radial-basis-function networks,'' \emph{Neural Networks}, vol.~14, no.~4, pp.
  439--458, 2001.

\bibitem{horn1990hadamard}
R.~A. Horn, ``The {H}adamard product,'' in \emph{Proc. Symp. Appl. Math},
  vol.~40, 1990, pp. 87--169.

\bibitem{destexhe2004plasticity}
A.~Destexhe and E.~Marder, ``Plasticity in single neuron and circuit
  computations,'' \emph{Nature}, vol. 431, no. 7010, pp. 789--795, 2004.

\bibitem{le1990handwritten}
Y.~LeCun, B.~Boser, J.~S. Denker, D.~Henderson, R.~E. Howard, W.~Hubbard, and
  L.~D. Jackel, ``Handwritten digit recognition with a back-propagation
  network,'' in \emph{Advances in Neural Information Processing Systems}, 1990.

\end{thebibliography}


\end{document}